\documentclass{article}
\usepackage[preprint]{neurips_2026}
\usepackage[utf8]{inputenc}
\usepackage[T1]{fontenc}
\usepackage{hyperref}
\usepackage{url}
\usepackage{booktabs}
\usepackage{amsfonts}
\usepackage{nicefrac}
\usepackage{microtype}
\usepackage{graphicx}
\usepackage{amsmath}
\usepackage{dsfont}
\newcommand{\mathbbm}[1]{\mathds{#1}}
\usepackage{multirow}
\usepackage[table]{xcolor}   
\usepackage{wrapfig}
\usepackage{enumitem}
\usepackage{algorithm}
\usepackage{algorithmic}
\usepackage[most]{tcolorbox}
\usepackage{titletoc}

\definecolor{accentR}{RGB}{212,120,92}     % terracotta (reasoning prompts)
\definecolor{accentB}{RGB}{87,133,168}     % blue (judge prompt)
\definecolor{pbg}{RGB}{253,251,248}        % warm off-white background

\newtcolorbox{reasoningprompt}[1]{
  enhanced, breakable,
  colback=pbg, colframe=pbg,
  boxrule=0pt, arc=0pt, outer arc=0pt,
  borderline west={2.5pt}{0pt}{accentR},
  left=8pt, right=6pt, top=5pt, bottom=5pt,
  fonttitle=\sffamily\bfseries\small\color{accentR},
  title={#1},
  attach title to upper={\par\smallskip},
  before skip=6pt, after skip=6pt
}

\newtcolorbox{judgeprompt}[1]{
  enhanced, breakable,
  colback=pbg, colframe=pbg,
  boxrule=0pt, arc=0pt, outer arc=0pt,
  borderline west={2.5pt}{0pt}{accentB},
  left=8pt, right=6pt, top=5pt, bottom=5pt,
  fonttitle=\sffamily\bfseries\small\color{accentB},
  title={#1},
  attach title to upper={\par\smallskip},
  before skip=6pt, after skip=6pt
}

\newcommand{\modelname}{BAR}

\title{Cite What You Explore: Budget-Aware LLM Reasoning over Medical KGs with Verifiable Evidence}

\author{
  Chen Chen$^{\dagger}$, 
  Dongjie Wang$^{\dagger}$,
  Mei Liu$^{\ddagger}$,
  Zijun Yao$^{\dagger}$\thanks{Corresponding author.}\\[6pt]
  $^{\dagger}$Electrical Engineering and Computer Science, University of Kansas, USA\\
  \texttt{\{chenchen, wangdongjie, zyao\}@ku.edu}\\[3pt]
  $^{\ddagger}$Health Outcomes and Biomedical Informatics, University of Florida, USA\\
  \texttt{mei.liu@ufl.edu}\\[3pt]
}

\begin{document}

\maketitle

\begin{abstract}
Post-discharge risk prediction from electronic health records (EHRs) is difficult because many dependencies that link discharge-time observations to downstream complications, such as comorbidity cascades and drug-disease interactions, are absent from the record.
External medical knowledge graphs (KGs) can supply these missing dependencies, but tracing them in a clinical setting demands three properties.
KG exploration must remain cost-bounded, retrieved evidence must be differentiated by source quality, and the resulting rationale must be citable for review.
Large language models (LLMs) can plan and verify over structured evidence, making them natural candidates for KG reasoning, but existing LLM-based methods do not satisfy these three properties jointly.
In this paper, we propose \textbf{\modelname{}}, a \underline{\textbf{B}}udget-\underline{\textbf{A}}ware LLM \underline{\textbf{R}}easoning framework over medical KGs with three contributions.
First, \modelname{} refines the raw KG into disease-specific evidence graphs whose edges carry support scores and provenance records, turning the KG into a quality-annotated reasoning space rather than a static feature source.
Second, an LLM reasons over this graph through a plan-navigate-verify loop that decomposes the question into steps, retrieves evidence under a patient-specific budget, and revises when verification fails.
Third, a reasoning policy is trained with a reward that compares predictions with and without acquired evidence, combined with acquisition cost and citation-integrity terms.
Across 8 diseases and 3 prediction horizons on MIMIC-III and MIMIC-IV, \modelname{} achieves higher performance and citation precision than strong baselines while using fewer LLM tokens.
Its gain over EHR-only prediction grows at longer horizons, and its advantage carries over to an external benchmark.
\end{abstract}

\section{Introduction}
\label{sec:introduction}

Electronic health records (EHRs) have enabled a wide range of post-discharge risk prediction tasks~\citep{rajkomar2018scalable, harutyunyan2019multitask}, where the goal is to assess at discharge whether a specific complication will first appear within a future time window. This task is challenging because the pathways that would connect discharge-time observations to downstream complications, including comorbidity cascades, drug-disease interactions, and disease progressions, are rarely captured in the record. External medical knowledge graphs (KGs)~\citep{chandak2023building} can supply the relational structure needed to recover these dependencies, but tracing them typically requires multi-hop exploration~\citep{zhang2021gmh}, since a single retrieval rarely captures the chain of intermediate concepts linking a discharge-time observation to its downstream complication.

Existing methods do not yet exploit this structure effectively. EHR-only models~\citep{choi2016retain, gao2020stagenet, rasmy2021med} capture statistical associations but miss risk factors that emerge only through pathways whose intermediate concepts the EHR does not record.
Static KG methods~\citep{choi2017gram, lu2021collaborative, jiang2024graphcare, tan2025boxlm} add relational knowledge but extract embeddings once before prediction, with no mechanism to follow patient-specific pathways at inference time.
What is missing is an agent that can decide which pathway to trace next and when the evidence is sufficient to stop for a given patient. Large language models (LLMs) are a natural fit~\citep{singhal2023large, kweon2024ehrnoteqa} because they can decompose a clinical scenario into hypothesis-driven steps and query the KG for evidence at each step, conditioning each retrieval on what previous steps have revealed.

Recent Graph4LLM approaches~\citep{zhao2026vog, ma2025thinkongraph, gao2025graph, chen2024planongraph} advance in this direction by allowing iterative reasoning over KGs, and medical KG methods such as KARE~\citep{jiang2025reasoningenhanced} retrieve summarized graph communities for each patient. However, three requirements for clinical use remain unmet. First, with no cost constraint, exploration continues well past the point of diminishing returns, even on patients whose discharge record alone already supports a confident prediction.
Second, treating all retrieved edges uniformly, regardless of how well-supported they are by underlying evidence, leaves the LLM unable to prioritize the most reliable pathways.
Third, the reasoning trace does not preserve a verifiable link between predictions and the provenance of supporting edges, so it cannot be audited.

\begin{figure}[t]
\centering
\includegraphics[width=\columnwidth]{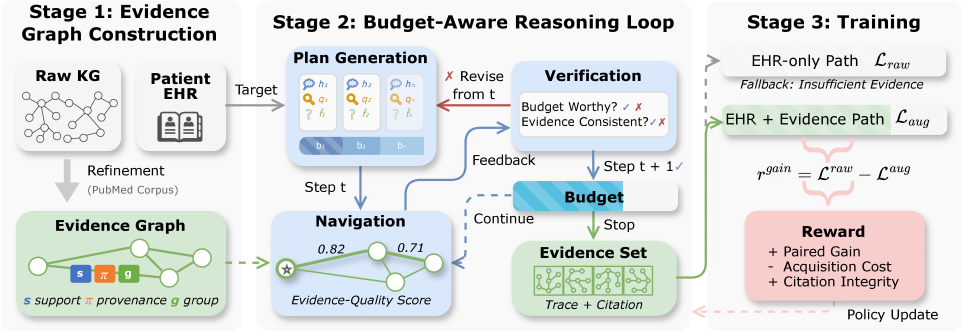}
\caption{
Overview of \modelname{}.
\textbf{Stage~1:} The raw KG is refined into a disease-specific evidence graph whose edges carry support scored against curated sources and a fixed PubMed corpus, provenance, and deduplication metadata.
\textbf{Stage~2:} The LLM generates a budget-aware reasoning plan, navigates the evidence graph using a patient-conditioned quality score, and verifies each step for consistency and the cost-quality trade-off, revising as needed.
\textbf{Stage~3:} The reasoning policy is trained via paired gain, acquisition cost, and citation-integrity reward.
}
\label{fig:framework}
\vspace{-10pt}
\end{figure}

To address these three requirements, we give the LLM and the KG complementary roles. The LLM generates and revises pathway hypotheses, while the KG constrains which pathways are structurally grounded and supplies the traceable evidence needed to verify them under bounded cost.
We develop this idea in \modelname{}\footnote{Code is available at \url{https://github.com/ChenC2002/BAR}.} (Figure~\ref{fig:framework}), a framework for budget-aware LLM reasoning over medical KGs. Our contributions are threefold. 
First, we introduce evidence graphs that refine the raw KG into disease-specific subgraphs whose edges carry support scores, provenance records, and equivalence-group ids that control redundancy, turning the KG into a quality-annotated reasoning space the LLM explores at inference time, rather than a feature source whose embeddings are extracted once before prediction.
Second, we propose a budget-aware reasoning loop that allocates exploration according to the training-set percentile of EHR-only prediction uncertainty, in which the LLM produces a clinical plan, navigates the evidence graph under this patient-adaptive budget, and verifies each step against both source support and the patient's evolving state, with revision when evidence falls short.
Third, we design a training signal whose primary term is a paired gain comparing predictions with and without acquired evidence, with auxiliary terms for acquisition cost and citation integrity, so the policy learns to acquire external evidence only when it improves prediction beyond what the discharge record alone provides.
Experiments on MIMIC-III~\citep{johnson2016mimic} and MIMIC-IV~\citep{johnson2023mimic} show that \modelname{} achieves higher AUROC and AUPRC than the compared baselines with fewer acquisition attempts and LLM tokens, and that a larger share of its citations is judged relevant.
Its relative gain over EHR-only prediction increases at longer horizons, and its advantage carries over to the external EHRSHOT benchmark~\citep{wornow2023ehrshot}.

\section{Methodology}
\label{sec:method}

\subsection{Problem Setup}
\label{sec:overview}
We predict the first recorded diagnosis of a target disease during a subsequent admission in the available EHR. Each eligible admission defines an instance for each target disease for which it meets the eligibility criteria, and discharge is the prediction time.
The index $i$ denotes an index admission and its associated patient history, with discharge time $\tau_i$.
The model input available at discharge has two parts, the diagnosis codes recorded during the index admission,
$X_i = \{c_{i,1}, \ldots, c_{i,n_i}\}$, where each $c_{i,j}$ is an ICD code, and a pre-discharge history $H_i$ that aggregates diagnosis codes from admissions strictly before $\tau_i$.

Let $\mathcal{D}$ denote a predefined set of target diseases and $\mathcal{T}$ denote a set of prediction horizons in days.
For index admission $i$, target disease $d \in \mathcal{D}$, and horizon $h \in \mathcal{T}$, the binary label is
\begin{equation}
y_{i,d}^{(h)}
=
\mathbbm{1}\!\left[
\exists\, t \in (\tau_i,\, \tau_i + h]
\text{ such that } d \text{ is first recorded at time } t
\right].
\label{eq:label}
\end{equation}
To focus on the first recorded post-discharge occurrence, we restrict to instances where $d$ has not appeared in either $H_i$ or $X_i$, and require a positive outcome to be documented during a subsequent admission.
First occurrence is defined from the available recorded history. Appendix~\ref{app:dataset} specifies the outcome timestamp and reports admission-level counts.
A zero label indicates no qualifying diagnosis within the prediction window, including instances with no recorded subsequent admission.
A patient may contribute multiple eligible index admissions, and one subsequent disease record can yield positive labels for multiple earlier eligible indices whose prediction windows include it.
The model outputs a risk score $\hat{y}_{i,d}^{(h)} \in [0,1]$ together with a cited evidence trace whose supporting edges can be inspected through stored provenance.

We represent an external biomedical knowledge graph such as PrimeKG~\citep{chandak2023building} as $\mathcal{G} = (\mathcal{V}, \mathcal{E})$.
For each target disease $d$, we construct a refined evidence graph $G_d = (V_d, E_d)$ from $\mathcal{G}$.
Each edge $e = (u, r, v) \in E_d$ is a typed triple with endpoint concepts $u, v \in V_d$ and relation type $r$.
Every retained edge carries per-edge metadata defined during evidence graph construction (Section~\ref{sec:support}).

We encode the patient's discharge-time EHR into a representation $\mathbf{z}_i^{\mathrm{ehr}}$ that summarizes $X_i$ and $H_i$.
To connect discharge-time observations to the evidence graph, we use a fixed linker $\operatorname{link}(\cdot)$ that maps ICD codes to concepts in $\mathcal{V}$.
For patient $i$ and target disease $d$, the anchor set is
$A_{i,d} = \{\operatorname{link}(c) : c \in X_i,\ \operatorname{link}(c) \in V_d\}$.
All graph interaction during reasoning is performed under a patient-specific budget $B_i$.

\subsection{Refined Evidence Graph}
\label{sec:evidence_graph}

This stage converts the raw knowledge graph into a disease-specific reasoning space that retains only supported, traceable, and low-redundancy edges.
Each graph is built per disease and shared across horizons, which enter later through the target embedding and the prediction head (Sections~\ref{sec:quality_score} and~\ref{sec:predictor}), so different horizons can follow different paths through the same graph.

\subsubsection{Candidate Region}
\label{sec:candidate_region}

Starting from disease node $d$, we collect all edges within $K$ hops, where $K$ is a fixed hop limit.
Let $\mathcal{E}_{\le K}(d)$ denote the set of edges in $\mathcal{E}$ whose endpoints are both within $K$ hops of $d$.
For reachability, we treat the graph as undirected, while each retained edge keeps its original relation type and direction.

We then restrict this region to edges connected to discharge-time observations.
Let $\mathcal{A}_d$ denote the set of frequent anchor concepts for disease $d$, obtained by applying $\operatorname{link}(\cdot)$ to frequent discharge-time codes in the training split.
We keep only edges that lie on at least one path from an anchor to the disease node:
\begin{equation}
E_d^{\mathrm{cand}}
=
\left\{
e \in \mathcal{E}_{\le K}(d)
\ \middle|\
\exists\, a \in \mathcal{A}_d
\text{ such that } e \text{ lies on a path from } a \text{ to } d
\right\}.
\label{eq:candidate_region}
\end{equation}

To reduce hub-driven noise, we apply a fixed degree threshold $d_{\max}$.
For any node whose degree exceeds $d_{\max}$ in the candidate region, we keep at most $M$ incident edges ranked by a fixed prior score $\tilde{s}(e) \in [0,1]$ that is available before support scoring.

\subsubsection{Support, Provenance, and Deduplication}
\label{sec:support}

For each candidate edge $e = (u, r, v) \in E_d^{\mathrm{cand}}$, we retrieve supporting sources and store their identifiers as the provenance record $\pi(e)$.
The sources are curated records from PrimeKG's constituent resources~\citep{chandak2023building} and abstracts from a fixed PubMed corpus.
The provenance record identifies the source and its release or publication date. Appendix~\ref{app:source_versions} specifies a fixed-snapshot construction and its temporal scope.
A fixed support scorer assigns $s(e) \in [0,1]$ using only the sources in $\pi(e)$.
The score is 0.4 times a curated term, which counts the curated resources that attest the edge, plus 0.6 times a text term, which measures the PubMedBERT~\citep{gu2021domain} similarity between the edge and its top three BM25-retrieved abstracts (Appendix~\ref{app:knowledge_preparation}).
We retain only edges with $s(e) \ge \tau_s$, yielding the filtered set $E_d^{\mathrm{keep}} \subseteq E_d^{\mathrm{cand}}$.
With $\tau_s = 0.3$, an edge attested by three curated resources is retained without textual support, while an edge with fewer attestations needs some literature support.

To reduce redundancy, we group semantically equivalent edges in $E_d^{\mathrm{keep}}$ and assign each edge an equivalence-group id $g(e)$.
Two edges are grouped only if they express the same medical fact with the same direction and comparable specificity.
Each group keeps one representative edge for citation, while all member edges are retained for traceability.

We also store a lightweight expansion guide $I_d$ for frequent anchors.
For each anchor $a \in \mathcal{A}_d$, the guide keeps a short ranked list of adjacent edge identifiers in $E_d$.
The final evidence graph is $G_d = (V_d, E_d)$, where $E_d = E_d^{\mathrm{keep}}$.

\subsection{Budget-Aware Reasoning Loop}
\label{sec:reasoning_loop}

Given a patient, a target disease, and a prediction horizon, this stage runs an LLM reasoning loop over $G_d$ under budget $B$.

\subsubsection{Evidence-Quality Score}
\label{sec:quality_score}

We first define a patient-conditioned quality score used throughout the loop.
Because the same discharge record may be relevant to different diseases and horizons in different ways, we condition the patient state on the prediction target.
For each target pair $(d,h)$, the initial patient state is
$\mathbf{p}_{i,d,h} = f_{\mathrm{state}}([\mathbf{z}_i^{\mathrm{ehr}};\, \mathbf{u}(d,h)])$,
where $f_{\mathrm{state}}(\cdot)$ is a learnable projection, $\mathbf{u}(d,h) \in \mathbb{R}^m$ is a learnable target embedding, and $[\cdot\,;\,\cdot]$ denotes concatenation.
After a step passes verification, the patient state updates via the  GRU-based~\citep{cho2014learning} transition $\mathbf{p}_{i,d,h}^{(t)} = \mathrm{GRU}(\mathbf{p}_{i,d,h}^{(t-1)},\, \bar{\mathbf{v}}^{(t)})$, where $\bar{\mathbf{v}}^{(t)}$ is the mean representation of edges selected at step $t$.
Failed steps leave the patient state unchanged.
All parameters of the reasoning policy are shared across diseases and horizons.
Only the lightweight target embedding $\mathbf{u}(d,h)$ and the prediction heads are task-specific.

Each concept node $x \in V_d$ has a learnable embedding $\mathbf{e}_x \in \mathbb{R}^m$.
For a candidate edge $e = (u, r, v) \in E_d$ under patient state $\mathbf{p}$, we define the evidence-quality score as
\begin{equation}
\phi(e;\, \mathbf{p})
=
s(e)\;\sigma\!\left(
\beta \cdot
\max\!\left\{
\mathbf{p} \cdot \mathbf{e}_u,\;
\mathbf{p} \cdot \mathbf{e}_v
\right\}
\right),
\label{eq:quality_score}
\end{equation}

where $\sigma(\cdot)$ is the sigmoid function and $\beta > 0$ is a fixed temperature that controls the sharpness of patient-relevance gating without adding a learnable scaling parameter.
We write $\phi_t(e)$ for the score under the step-$t$ state and $\phi^\star(e)$ for the score under the terminal state.
The multiplicative form acts as a soft AND gate, requiring an edge to be both well-supported and patient-relevant to receive a high score.
Appendix~\ref{app:quality_analysis} analyzes the joint effect of support and patient relevance on selection rates.

\subsubsection{Budget and Cost}
\label{sec:budget}

When the discharge record already supports a confident prediction, additional evidence acquisition adds cost without improving accuracy, so we allocate larger budgets to uncertain EHR-only predictions from stage~1.

For target $(d,h)$, let $\hat{y}_{i,d,h}^{\mathrm{ehr}}$ be the frozen EHR-only score.
We define $\tilde{u}_i=1-|2\hat{y}_{i,d,h}^{\mathrm{ehr}}-1|$ and use its training-set percentile $u_i=F_{d,h}^{\mathrm{train}}(\tilde{u}_i)$.
Uncertainty is highest near a predicted risk of 0.5.
The CDF $F_{d,h}^{\mathrm{train}}$ is estimated separately for each disease and horizon and held fixed for validation and testing.
We omit the fixed target from uncertainty and budget subscripts.
The patient-specific budget is
\begin{equation}
B_i = B_{\min} + (B_{\max} - B_{\min}) \cdot u_i,
\label{eq:adaptive_budget}
\end{equation}
where $B_{\min} = 2$ allows an initial acquisition attempt and $B_{\max} = 12$ caps total exploration.

Within this budget, each acquisition step $t$ incurs a cost determined
by three binary indicators.
The indicators $\delta_t^q$ and $\delta_t^{\mathrm{exp}}$ record whether
\textsc{QueryGuide} or \textsc{QueryExpand} is executed.
These query types are mutually exclusive.
The indicator $\delta_t^{\mathrm{sel}}$ records whether the selection
operation is executed.
An executed selection operation is charged even when it returns no
edges, and subsequent verification failure does not refund the cost.
The per-step cost is
\begin{equation}
c_t
=
\alpha_q \delta_t^{q}
+
\alpha_{\mathrm{exp}} \delta_t^{\mathrm{exp}}
+
\alpha_{\mathrm{sel}} \delta_t^{\mathrm{sel}},
\label{eq:step_cost}
\end{equation}
where $\alpha_q, \alpha_{\mathrm{exp}}, \alpha_{\mathrm{sel}} \ge 0$ are fixed weights.
The cumulative cost up to step $t$ is $C_t = \sum_{j=1}^{t} c_j$.

\subsubsection{Plan Generation}
\label{sec:plan_generation}

The LLM produces an initial reasoning plan that decomposes the clinical question into a sequence of evidence-gathering steps.
Given patient EHR $(X_i, H_i)$, target disease $d$, horizon $h$, and anchor set $A_{i,d}$, we prompt the LLM to produce a plan $P = [s_1, s_2, \ldots, s_n]$, where $n$ is determined by the LLM.
Each step $s_t = (h_t, q_t, \hat{f}_t, b_t)$ contains a clinical hypothesis $h_t$, a query intent $q_t$, an expected evidence description $\hat{f}_t$, and a planned budget fraction $b_t \in (0,1]$, with $\sum_t b_t \leq 1$.
The fractions guide the LLM's planning and revision.
Execution enforces the episode-level budget by requiring each acquisition cost to fit within the remaining budget $B_i-C_{t-1}$.

\subsubsection{Plan-Guided Graph Navigation}
\label{sec:navigation}

At plan step $t$, the query intent $q_t$ guides retrieval from $G_d$ through two query types.
\textsc{QueryGuide}$(a)$ starts from an anchor $a \in A_{i,d}$ and returns candidates from the expansion guide $I_d$ when $a \in \mathcal{A}_d$, falling back to direct lookup of incident edges otherwise.
\textsc{QueryExpand}$(x,r)$ expands a frontier concept $x$ along relation type $r$ and returns incident edges in $E_d$.
The first step of an episode uses \textsc{QueryGuide} to ground the reasoning in the patient's discharge-time observations, while subsequent steps use \textsc{QueryExpand} to follow the pathway outward from concepts discovered in earlier steps.

After a query returns candidates, we score each with $\phi_t(e)$, keep only the top $K_{\mathrm{pre}}$, and remove any candidate whose equivalence-group id has already been selected.
The filtered candidates are presented to the LLM as graph feedback $f_t$, consisting of typed triples with their support scores.

The LLM then selects edges to retain.
Each selected edge must satisfy three executable constraints requiring that the edge exists in $E_d$, its support score satisfies $s(e) \ge \tau_s$, and it belongs to an equivalence group that has not been previously selected.
Selected edges are added to the current evidence set $E_{i,d,h}^{(t)}$ only if the step passes verification.

Coherence comes from three inductive biases.
The patient state aggregates previously selected evidence through recurrent updates, expansion starts from anchors or endpoints of accepted evidence, and equivalence-group exclusion prevents redundant relations.

\subsubsection{Verification and Revision}
\label{sec:verification}

After each navigation step, the LLM compares the graph feedback $f_t$ against its expected evidence $\hat{f}_t$ and triggers a revision when the feedback contradicts or fails to support its hypothesis.
Verification proceeds in two stages.
First, deterministic admissibility checks enforce hard constraints on support, equivalence-group uniqueness, and minimum evidence quality.
Once these checks pass, the LLM verifies semantically whether the evidence coherently supports the clinical hypothesis.

On failure, the evidence set and patient state remain unchanged. 
If enough budget remains, the LLM regenerates the remaining plan from the failed step onward, keeping verified earlier steps unchanged. 
The revision prompt includes the current feedback, the accumulated reasoning context, and the remaining budget $B_i-C_t$, allowing the LLM to adapt plan length to actual consumption.

% [S-1] Removed 'The failed attempt remains charged.' from the end of 2.3.5. Section 2.3.2 already states that failed attempts are charged.
\subsubsection{Stopping}
\label{sec:stopping}

To detect when further exploration yields diminishing returns, we track the set-level utility
\begin{equation}
U_t
=
\frac{1}{\max\{1,\, |E_{i,d,h}^{(t)}|\}}
\sum_{e \in E_{i,d,h}^{(t)}} \phi_t(e),
\label{eq:utility}
\end{equation}
where all selected edges are rescored under the current patient state at step $t$.

The loop terminates when the remaining budget $B_i - C_t$ cannot support another feasible action, when $U_t$ improves by less than $\epsilon_{\mathrm{stop}}$ for $P_{\mathrm{stop}}$ consecutive steps, or when the evidence set reaches a size cap $K_E$.
If the final evidence set is too small ($|E_{i,d,h}| < K_{\min}$) or too weak ($U_{\mathrm{final}} < \tau_{\mathrm{fb}}$), the model falls back to the EHR-only prediction.

\subsection{Prediction and Training}
\label{sec:training}

\subsubsection{Predictor}
\label{sec:predictor}

For episodes that pass the fallback thresholds, we summarize the final evidence set $E_{i,d,h}$ using score-weighted pooling.
Let $\mathbf{v}(e) \in \mathbb{R}^p$ denote the learned representation of edge $e$.
The evidence representation is
\begin{equation}
\mathbf{z}_{i,d,h}^{\mathrm{kg}}
=
\sum_{e \in E_{i,d,h}} w(e)\,\mathbf{v}(e),
\qquad
w(e)
=
\frac{\phi^\star(e)}
{\max\!\left(
\epsilon,\,
\sum_{e' \in E_{i,d,h}} \phi^\star(e')
\right)},
\label{eq:evidence_pool}
\end{equation}
where $\epsilon > 0$ is a small constant. 
On fallback, the model returns the frozen EHR-only score
$\hat{y}_{i,d,h}^{\mathrm{ehr}}$ and an empty citation list.
Otherwise, the predictor combines $\mathbf{z}_i^{\mathrm{ehr}}$
from Section~\ref{sec:overview} with the evidence representation.
The final risk score is
\begin{equation}
\hat{y}_{i,d}^{(h)}
=
\sigma\!\left(
\mathbf{t}_{d,h}^{\top}
f\!\left(
[\mathbf{z}_i^{\mathrm{ehr}};\,
 \mathbf{z}_{i,d,h}^{\mathrm{kg}}]
\right)
+ b_{d,h}
\right),
\label{eq:prediction}
\end{equation}
where $f(\cdot)$ is a shared trainable fusion function with output in $\mathbb{R}^p$.
Each disease-horizon pair has a prediction head with parameters $\mathbf{t}_{d,h}\in\mathbb{R}^p$ and $b_{d,h}\in\mathbb{R}$.
The LLM therefore controls evidence acquisition, and its plans, hypotheses, budget fractions, and verification decisions influence the risk score only through the edges that enter $E_{i,d,h}$.
The citation list is formed by sorting edges in $E_{i,d,h}$ by $\phi^\star(e)$, deduplicating by $g(e)$, and keeping at most $K_C$ edges, each retaining its support score, provenance record, and equivalence-group membership for inspection.

\subsubsection{Reward}
\label{sec:reward}

We compare predictions with and without acquired evidence under the same model snapshot.
Let $\mathcal{L}_{i,d,h}^{\mathrm{raw}}$ denote the binary cross-entropy loss when the evidence representation is set to zero, and let $\mathcal{L}_{i,d,h}^{\mathrm{aug}}$ denote the loss when $E_{i,d,h}$ is used.
The paired gain is
\begin{equation}
r_{i,d,h}^{\mathrm{gain}}
=
\mathcal{L}_{i,d,h}^{\mathrm{raw}}
-
\mathcal{L}_{i,d,h}^{\mathrm{aug}}.
\label{eq:paired_gain}
\end{equation}
Both losses are computed on the same batch with dropout disabled.

The citation-integrity reward $r_{i,d,h}^{\mathrm{qual}}$ is 1 when
the final citation list is non-empty and passes all integrity checks,
and 0 otherwise.
Every cited edge must belong to $E_d$, satisfy $s(e)\geq\tau_s$,
and have a non-empty provenance record.
The list must contain at most $K_C$ edges with distinct
equivalence-group identifiers.
Fallback episodes receive zero because they return no citations.
Patient-specific relevance is evaluated separately by the citation judge.
The full episode reward is
\begin{equation}
r_{i,d,h}
=
\lambda_{\mathrm{gain}}\, r_{i,d,h}^{\mathrm{gain}}
-
\lambda_{\mathrm{cost}}\, \frac{C_{i,d,h}}{B_{\max}}
+
\lambda_{\mathrm{qual}}\, r_{i,d,h}^{\mathrm{qual}},
\label{eq:reward}
\end{equation}
where $C_{i,d,h} = \sum_t c_t$ is the total acquisition cost, $B_{\max}$ is the maximum budget and serves as a fixed normalizer, and $\lambda_{\mathrm{gain}}, \lambda_{\mathrm{cost}}, \lambda_{\mathrm{qual}} \ge 0$ are fixed weights.

\subsubsection{Training Procedure}
\label{sec:training_procedure}

Training proceeds in three stages, detailed in Appendix~\ref{app:impl}.
In the first stage, we train an EHR-only baseline by setting $\mathbf{z}_{i,d,h}^{\mathrm{kg}} = \mathbf{0}$ for all instances and optimizing the predictor with binary cross-entropy loss.
This baseline is frozen for budget allocation and fallback. 
The paired losses in Eq.~\eqref{eq:paired_gain} use the current predictor with and without acquired evidence.

In the second stage, the reasoning loop runs with a fixed exploration policy that executes a small number of queries and selections per episode, without policy optimization.
The predictor is updated with supervised loss while conditioning on the acquired evidence.

In the third stage, we update the reasoning policy using REINFORCE~\citep{zhang2021sample} with the episode reward in Eq.~\eqref{eq:reward}, while continuing supervised predictor updates.
Appendix~\ref{app:routing_policy} specifies the routing distribution, trainable parameters, and update, and Appendix~\ref{app:impl} reports the optimization settings.

\section{Experiments}
\label{sec:experiments}
 
\subsection{Setup}
\label{sec:setup}

\paragraph{Datasets.}
We evaluate on MIMIC-III~\cite{johnson2016mimic, harutyunyan2019multitask} and MIMIC-IV~\cite{johnson2023mimic}.
We define 8 target diseases spanning comorbidity cascades (chronic kidney disease, heart failure, COPD exacerbation), drug-disease interactions (hyperkalemia, hypoglycemia), and disease-progression pathways (ischemic stroke, sepsis, deep vein thrombosis).
Prediction horizons are $\mathcal{T} = \{30, 90, 365\}$ days, which correspond to transitional care, follow-up planning for chronic disease, and longer-term risk stratification.
We use a 70/10/20 split based on recorded discharge dates, with all admissions from a patient assigned to one partition.
For an external site, we use the six one-year new-diagnosis tasks of EHRSHOT~\cite{wornow2023ehrshot}.
Appendix~\ref{app:dataset} details the cohorts.

\paragraph{Knowledge Graph.}
We use PrimeKG~\cite{chandak2023building}, which integrates 20 biomedical resources into approximately 129K nodes and 4.0M edges across 30 relation types.
After disease-specific refinement (Section~\ref{sec:evidence_graph}), the average evidence graph $G_d$ contains approximately 5.2K nodes and 14.8K edges, with 45\% of candidate edges removed by the support threshold and about 12\% grouped as redundant (Appendix~\ref{app:dataset}).

\paragraph{Baselines.}
We compare against three categories of methods.
EHR-only models use no external knowledge and include RETAIN~\cite{choi2016retain}, StageNet~\cite{gao2020stagenet}, and Med-BERT~\cite{rasmy2021med}.
Static KG methods incorporate external knowledge as fixed features and include GRAM~\cite{choi2017gram} and GraphCare~\cite{jiang2024graphcare}.
Graph4LLM methods reason over the same evidence graph $G_d$ without budget constraints, source-quality scoring, or citation-integrity training, and include ToG 2.0~\cite{ma2025thinkongraph}, VoG~\cite{zhao2026vog}, and KARE~\cite{jiang2025reasoningenhanced}.
ToG-adapt does not verify retrieved edges, VoG-adapt verifies each step, and KARE-adapt retrieves community summaries.
All LLM-based methods use GPT-4o-mini~\cite{hurst2024gpt}, and we also report \modelname{} with Qwen3-8B~\cite{yang2025qwen3}.
Appendix~\ref{app:impl} describes the adaptations.

\paragraph{Metrics.}
We report AUROC and AUPRC for each disease-horizon pair and take the macro-average across all 24 tasks, with AUPRC as the primary metric because positive outcomes are uncommon.
Citation precision estimates the fraction of displayed citations that a GPT-4o judge labels relevant to the patient's risk pathway using a uniform sample of returned citations. On pairs where two annotators agree, the judge matches their label with a Cohen's $\kappa$ of 0.83 (Appendix~\ref{app:citation_eval}).
Evidence quality is the mean $\phi^\star$ of displayed citations, and both citation metrics exclude fallback episodes.
Budget utilization is the share of the assigned budget that is spent, a reasoning step is one acquisition attempt including failed ones, and efficiency is measured in LLM tokens per episode.
For clinical utility, we report the Brier score~\cite{brier1950verification}, the integrated calibration index~(ICI)~\cite{austin2019integrated}, and decision-curve net benefit~\cite{vickers2006decision} after recalibration on the validation split.
Table~\ref{tab:main} reports means and sample standard deviations over three independent training seeds on the same data split. Appendix~\ref{app:impl} gives the remaining settings.

\subsection{Main Results}
\label{sec:main_results}

\begin{table}[t]
\caption{Main results across eight diseases and three horizons over three independent training seeds. All metrics except steps are percentages. Bold marks \modelname{} with GPT-4o-mini, and underlining marks the strongest non-\modelname{} baseline for each prediction metric.}
\label{tab:main}
\centering
\renewcommand{\arraystretch}{1.2}
\resizebox{\columnwidth}{!}{%
\scriptsize
\setlength{\tabcolsep}{12pt}
\begin{tabular}{llcccc}
\toprule
& & \multicolumn{2}{c}{MIMIC-III} & \multicolumn{2}{c}{MIMIC-IV} \\
\cmidrule(lr){3-4}\cmidrule(lr){5-6}
Category & Method & AUROC & AUPRC & AUROC & AUPRC \\
\midrule
EHR-only & RETAIN & 71.31$_{\pm0.38}$ & 9.53$_{\pm0.21}$ & 70.46$_{\pm0.48}$ & 9.13$_{\pm0.49}$ \\
 & StageNet & 72.89$_{\pm0.36}$ & 10.63$_{\pm0.36}$ & 72.39$_{\pm0.26}$ & 9.66$_{\pm0.24}$ \\
 & Med-BERT & 74.46$_{\pm0.43}$ & 11.44$_{\pm0.25}$ & 73.97$_{\pm0.94}$ & 11.04$_{\pm0.44}$ \\
Static KG & GRAM & 75.10$_{\pm0.59}$ & 12.45$_{\pm0.22}$ & 75.38$_{\pm0.61}$ & 12.46$_{\pm0.30}$ \\
 & GraphCare & 77.34$_{\pm0.42}$ & 14.29$_{\pm0.37}$ & 76.88$_{\pm0.42}$ & 13.69$_{\pm0.29}$ \\
Graph4LLM 
& KARE-adapt & 78.49$_{\pm0.30}$ & 16.06$_{\pm0.28}$ & 77.77$_{\pm0.48}$ & \underline{15.88}$_{\pm0.17}$ \\
& ToG-adapt & 78.61$_{\pm0.53}$ & \underline{16.20}$_{\pm0.51}$ & 77.78$_{\pm0.32}$ & 14.56$_{\pm0.23}$ \\
& VoG-adapt & \underline{79.21}$_{\pm0.70}$ & 15.94$_{\pm0.29}$ & \underline{78.60}$_{\pm0.38}$ & 15.41$_{\pm0.45}$ \\
\rowcolor[gray]{0.90} \modelname{} (Ours) & Qwen3 & 80.43$_{\pm0.27}$ & 17.08$_{\pm0.87}$ & 79.23$_{\pm0.54}$ & 16.85$_{\pm0.62}$ \\
\rowcolor[gray]{0.90}  & \textbf{GPT-4o-mini} & \textbf{82.41}$_{\pm0.65}$ & \textbf{19.50}$_{\pm0.24}$ & \textbf{81.90}$_{\pm0.19}$ & \textbf{18.70}$_{\pm0.48}$ \\
\bottomrule
\end{tabular}}
\smallskip
\resizebox{\columnwidth}{!}{%
\scriptsize
\setlength{\tabcolsep}{5pt}
\begin{tabular}{lccccc}
\toprule
\multicolumn{6}{c}{\textbf{Reasoning Quality \& Efficiency} (MIMIC-III\,/\,MIMIC-IV)} \\
\midrule
Method & Citation Precision & Evidence Quality & Budget Utilization & Reasoning Steps & Fallback Rate \\
\midrule
ToG-adapt & 53.21$_{\pm0.72}$\,/\,50.21$_{\pm1.00}$ & 47.98$_{\pm0.73}$\,/\,46.22$_{\pm1.11}$ & --- & 6.65$_{\pm0.11}$\,/\,7.07$_{\pm0.12}$ & --- \\
VoG-adapt & 59.00$_{\pm1.45}$\,/\,56.63$_{\pm1.81}$ & 48.20$_{\pm1.01}$\,/\,45.92$_{\pm1.74}$ & --- & 5.57$_{\pm0.12}$\,/\,6.01$_{\pm0.15}$ & --- \\
\rowcolor[gray]{0.90} \modelname{} (Qwen3) & 72.04$_{\pm0.86}$\,/\,64.55$_{\pm0.99}$ & 56.78$_{\pm0.75}$\,/\,52.83$_{\pm1.22}$ & 69.72$_{\pm1.96}$\,/\,70.54$_{\pm1.32}$ & 4.38$_{\pm0.14}$\,/\,4.44$_{\pm0.09}$ & 22.26$_{\pm0.92}$\,/\,27.04$_{\pm2.34}$ \\
\rowcolor[gray]{0.90} \textbf{\modelname{} (GPT-4o-mini)} & \textbf{77.06}$_{\pm0.81}$\,/\,\textbf{73.64}$_{\pm1.19}$ & \textbf{64.84}$_{\pm1.74}$\,/\,\textbf{60.86}$_{\pm0.86}$ & \textbf{61.86}$_{\pm1.27}$\,/\,\textbf{65.06}$_{\pm1.35}$ & \textbf{3.83}$_{\pm0.09}$\,/\,\textbf{4.05}$_{\pm0.09}$ & \textbf{15.54}$_{\pm0.34}$\,/\,\textbf{19.75}$_{\pm0.82}$ \\
\bottomrule
\end{tabular}}
\end{table}

\modelname{} achieves the highest AUROC and AUPRC on both datasets (Table~\ref{tab:main}), with an AUPRC advantage of about three percentage points over the strongest baseline, which is ToG-adapt on MIMIC-III and KARE-adapt on MIMIC-IV.
\modelname{} remains ahead with either LLM. Qwen3-8B gives lower prediction and citation quality than GPT-4o-mini but retains an advantage over the baselines.
The relative AUPRC gain over the best EHR-only baseline grows at longer horizons (Appendix~\ref{app:full_results}). External evidence may help capture relationships that are less apparent in the discharge record, although outcome prevalence also varies with the horizon.

\modelname{} combines higher citation precision with fewer acquisition attempts and fewer tokens.
It uses about two thirds of the tokens of ToG-adapt and half those of VoG-adapt, mainly because each step shows the LLM only $K_{\mathrm{pre}}$ filtered candidates.
Most episodes use several attempts, allowing the system to verify intermediate evidence and revise unsuccessful paths. ToG-adapt still makes fewer LLM calls because it omits verification (Table~\ref{tab:latency}).

\subsection{Ablation Study}
\label{sec:ablation}
 
\begin{wrapfigure}{r}{0.62\columnwidth}
\vspace{-10pt}
\centering
\includegraphics[width=0.62\columnwidth]{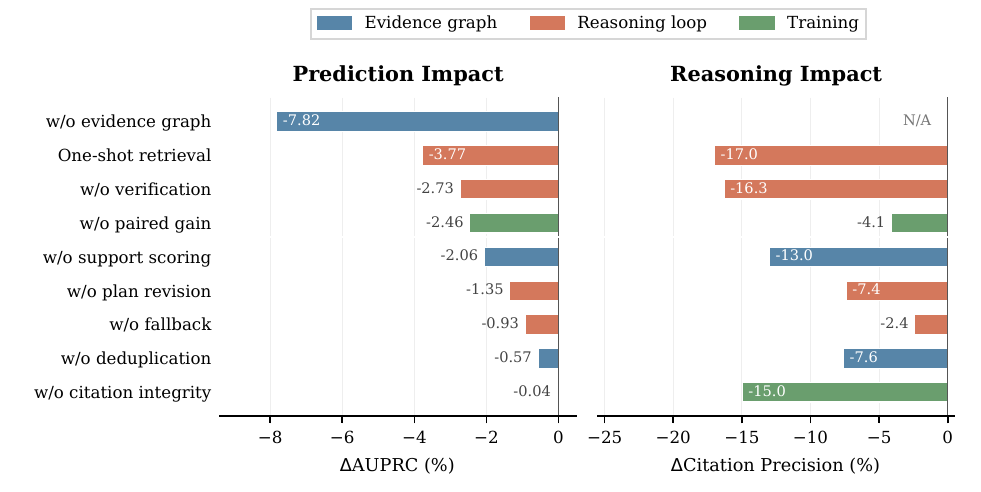}
\vspace{-10pt}
\caption{Ablation results from the full model, macro-averaged over tasks and datasets. Colors indicate framework stage.}
\label{fig:ablation}
\vspace{-10pt}
\end{wrapfigure}
 
Figure~\ref{fig:ablation} compares each component removal with the full model.
Removing the evidence graph causes the largest prediction loss, while several other components have a stronger effect on citation precision.
One-shot retrieval reduces both metrics, and removing verification lowers AUPRC more than removing plan revision.
The w/o paired gain variant continues stage~3 policy optimization with cost and citation-integrity rewards, while stage~2 uses uniform routing.
Removing paired gain substantially lowers AUPRC, whereas removing citation integrity leaves prediction nearly unchanged but substantially reduces citation precision.
At matched budgets, support-greedy navigation loses more AUPRC than removing verification, and random navigation falls below GraphCare (Appendix~\ref{app:navigation}).

\clearpage
\subsection{Analysis}
\label{sec:analysis}

\begin{wrapfigure}{r}{0.68\columnwidth}
\vspace{-15pt}
\centering
\includegraphics[width=0.68\columnwidth]{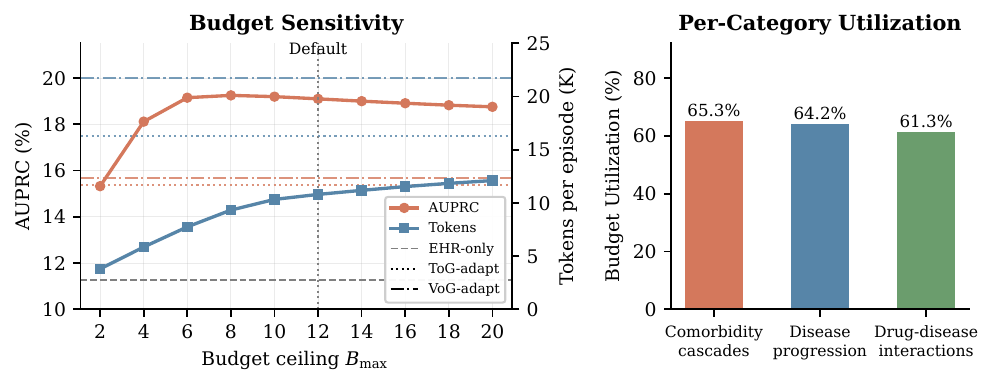}
\vspace{-15pt}
\caption{AUPRC and mean tokens per episode across budget ceilings, with ToG-adapt and VoG-adapt as reference lines (left), and per-category budget utilization (right), averaged over both datasets.}
\label{fig:budget}
\vspace{-5pt}
\end{wrapfigure}

\paragraph{Budget Sensitivity.}
AUPRC rises quickly at low budgets, peaks at $B_{\max}=8$ just below the validation-selected default, and declines slightly while token use keeps rising (Figure~\ref{fig:budget}, left).
From $B_{\max}=4$ upward, \modelname{} exceeds ToG-adapt and VoG-adapt in AUPRC, and it uses fewer tokens than both at every ceiling.
Comorbidity cascades use a larger share of their assigned budgets than drug-disease interactions (Figure~\ref{fig:budget}, right), and since every category follows the same allocation rule, this difference reflects how far the loop explores.
A plausible clinical reading is that a drug-disease interaction is often supported by a direct drug-effect relation, whereas a comorbidity cascade passes through intermediate conditions and organ systems in larger evidence graphs (Table~\ref{tab:kg}).
Drug-disease interactions also fall back more often (Table~\ref{tab:fallback_coverage}), which suggests the loop stops early when a short pathway is missing or weakly supported.

\paragraph{Robustness.}
\modelname{}'s relative advantage over VoG-adapt grows as fewer ICD codes are retained.
On held-out diseases, its relative decline is larger than GraphCare's but smaller than VoG-adapt's, while its absolute AUPRC remains higher than both (Appendix~\ref{app:robustness}).
The advantage also holds across graph-construction settings, with UMLS in place of PrimeKG, under reduced linker coverage, and on EHRSHOT, and it narrows but persists when laboratory values and medications are added.
Corrupting navigation candidates lowers AUPRC linearly while raising fallback (Appendices~\ref{app:robustness} and~\ref{app:sensitivity}).
 
\paragraph{Fallback Behavior.}
Fallback is more frequent on MIMIC-IV than on MIMIC-III (Table~\ref{tab:main}).
Removing it lowers overall AUPRC, which supports an evidence threshold before graph information enters the prediction, at the cost of no citations for the episodes that fall back (Appendix~\ref{app:full_results}).

\paragraph{Lead Time.}
Because a diagnosis recorded within days of discharge may reflect a condition already present but not yet coded, we remove index admissions whose target disease is first recorded within seven days of discharge and retrain every method.
AUPRC drops for every method by a similar fraction, and \modelname{} remains strongest on both datasets.
Its margin disappears at 30 days, where early events make up about a quarter of positives, and grows with the horizon, so external pathways add most for later outcomes (Appendix~\ref{app:leadtime_results}).
 
\paragraph{Clinical Utility.}
Raw scores of \modelname{}, KARE-adapt, and the stage-1 EHR-only model are similarly miscalibrated, and recalibration on the validation split lowers ICI about fourfold.
After recalibration, \modelname{} has the highest decision-curve net benefit at every threshold and horizon, and its relative lead grows as the threshold rises past the outcome prevalence (Appendix~\ref{app:utility}).

\subsection{Case Study}
 
\begin{figure}[t]
\vspace{-5pt}
\centering
\includegraphics[width=\columnwidth]{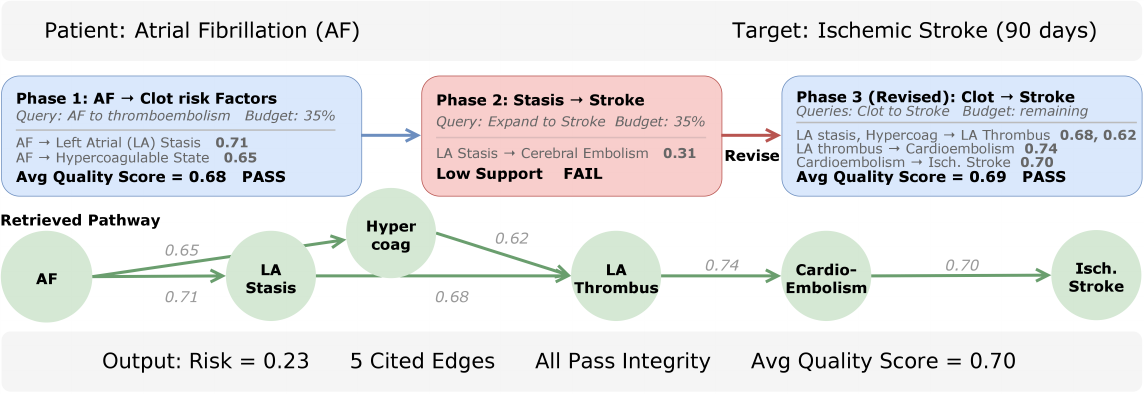}
\vspace{-10pt}
\caption{Reasoning trace for a patient with atrial fibrillation, from the initial plan through a failed step and its revision. Phase~3 groups at least three adjacency queries.}
\label{fig:case_study}
\vspace{-10pt}
\end{figure}

The patient in Figure~\ref{fig:case_study} has atrial fibrillation.
The initial query retrieves evidence for left atrial stasis and a
hypercoagulable state.
A proposed shortcut from stasis to cerebral embolism fails verification
because the edge has weak support.
The revised plan follows a pathway through left atrial thrombus and
cardioembolism to ischemic stroke, retaining the previously accepted evidence.
This expansion requires successive adjacency queries as new frontier
nodes become available.
Six edges are accepted, and the five highest-scoring edges form the
final citation list within $K_C=5$.

\section{Related Work}
\label{sec:related_work}

\paragraph{Clinical Prediction from EHRs.}
Visit-sequence models such as RETAIN~\citep{choi2016retain}, Dipole~\citep{ma2017dipole}, StageNet~\citep{gao2020stagenet}, GCT~\cite{choi2020learning}, AdaCare~\cite{ma2020adacare}, and pretrained transformers~\citep{rasmy2021med, li2020behrt} learn temporal patterns from coded diagnoses and procedures but have no mechanism to recover risk pathways that are absent from the record.
KG-augmented methods such as GRAM~\citep{choi2017gram}, CGL~\citep{lu2021collaborative}, GraphCare~\citep{jiang2024graphcare}, and BoxLM~\citep{tan2025boxlm} address this gap by incorporating relational knowledge from external graphs, yet they treat the graph as a source of embeddings extracted before prediction rather than as a reasoning space explored at inference time.
KARE~\citep{jiang2025reasoningenhanced} moves closer by organizing a multi-source graph into summarized communities and retrieving community summaries for each patient before an LLM reasons toward a prediction.
Because each summary condenses many relations, its predictions do not cite individual sourced edges, and its retrieval has no patient-specific acquisition budget.

\paragraph{LLM Reasoning over KGs.}
Structured multi-step reasoning~\cite{wei2022chain,yao2023tree} and retrieval-augmented generation~\cite{lewis2020retrieval} have enabled LLMs to iteratively plan, retrieve, and verify over knowledge graphs~\citep{ma2025thinkongraph,luo2024reasoning,chen2024planongraph,gao2025graph,zhao2026vog}.
These Graph4LLM methods assume an open-domain setting where exploration can run until the model chooses to stop, and all retrieved edges are treated uniformly.
Cost-sensitive feature acquisition~\citep{janisch2019classification} introduces budget constraints but operates on flat feature vectors without graph structure.
Adaptive retrieval~\citep{asai2023self,jiang2023active} decides when to retrieve based on expected utility but does not maintain source provenance across retrieval steps.
RL-based graph traversal~\citep{das2018go} learns exploration policies over graphs but does not differentiate edges by evidential backing.
\modelname{} addresses all three requirements jointly in a clinical prediction setting.

\section{Conclusion and Limitations}
\label{sec:conclusion}
 
We presented \modelname{}, a framework for budget-aware LLM reasoning over medical knowledge graphs for post-discharge risk prediction.
Across eight diseases and three prediction horizons on MIMIC-III and MIMIC-IV, \modelname{} achieves higher prediction and citation quality than the compared baselines with fewer LLM tokens, and its gain over EHR-only prediction grows at longer horizons and carries over to an external benchmark.
Its labels cover only diagnoses recorded within one health system and do not model censoring or competing risks, its raw scores need local recalibration, and the medical correctness of its full reasoning traces remains untested.
Prospective clinical validation is needed to assess whether source-backed rationales improve clinician trust in practice.
 
\begin{ack}
This work was supported by the National Science Foundation under Grant No.~OIA-2531881.
We also thank the anonymous reviewers for their valuable suggestions and feedback.
The authors declare no competing interests.
\end{ack}

\bibliographystyle{plain}
\bibliography{references}

\clearpage
\appendix
% Flush pending floats before each appendix section so tables stay in their section.
\makeatletter
\providecommand\FloatBarrier{\par\begingroup\let\@elt\relax
  \edef\@tempa{\@botlist\@deferlist\@dbldeferlist}%
  \ifx\@tempa\@empty\else\clearpage\fi\endgroup}
\makeatother
 
\startcontents[appendix]
\section*{Appendix Contents}
\begingroup
\hypersetup{hidelinks}
\printcontents[appendix]{l}{1}{\setcounter{tocdepth}{2}}
\endgroup
\vspace{1em}

\clearpage

\section{Dataset and Knowledge Graph Statistics}
\label{app:dataset}

\subsection{Prediction Cohorts}
\label{app:prediction_cohorts}

\paragraph{Index admissions.}
Candidate indices are adult hospital admissions with at least one linked
ICU stay, valid admission and discharge times, and discharge alive.
Hospital admissions are counted once, regardless of the number of linked
ICU stays or diagnosis records.
Prior history and subsequent outcomes include all hospital admissions
in each patient's history.
Hospital episodes do not overlap.
For each disease, we exclude indices with the target diagnosis in the
index admission or any earlier admission.
All remaining indices are retained, and the same set is used at every
prediction horizon.
A patient may contribute more than one index admission for a disease.

\paragraph{Outcome timing.}
Hospital diagnosis codes describe a stay and are assigned for billing at discharge~\citep{johnson2016mimic,johnson2023mimic}. We date an outcome by the discharge of the subsequent admission carrying the diagnosis. Let $a_j$ and $\tau_j$ denote admission $j$'s admission and discharge times, and let $p_j$ identify its patient. For an eligible index $i$ and disease $d$, the first recorded event time is
\begin{equation}
t_{i,d}^{\star}=\min\{\tau_j: p_j=p_i,\ a_j>\tau_i,\ j\text{ carries a diagnosis of }d\},
\label{eq:event_time}
\end{equation}
with $t_{i,d}^{\star}=+\infty$ when no qualifying admission exists. Ties are resolved by admission identifier. The label is $y_{i,d}^{(h)}=\mathbbm{1}[t_{i,d}^{\star}\leq\tau_i+h]$. This timestamp identifies the recorded diagnosis rather than biological onset. A zero label denotes no qualifying recorded diagnosis within the window, including cases without a recorded readmission.

\paragraph{Counting units.}
For the eligible index set $\mathcal I_d$, the cohort size, positive count, and positive rate are
\begin{equation}
N_d=|\mathcal I_d|,\qquad
P_d(h)=\sum_{i\in\mathcal I_d}y_{i,d}^{(h)},\qquad
\mathrm{Rate}_d(h)=100\,P_d(h)/N_d.
\label{eq:cohort_counts}
\end{equation}
Let $\mathcal I_0$ be the candidate index set before disease-specific exclusions, with $N_0=|\mathcal I_0|$. The count $F_0(h)$ includes candidate indices with at least one hospital admission that starts after the index discharge and ends within $h$ days. Applying the same rule to $\mathcal I_d$ gives $F_d(h)$. Each index is counted once, even with several returns. These counts satisfy $P_d(h)\leq F_d(h)\leq N_d$ and $F_d(h)\leq F_0(h)$. For $F_d(h)>0$, $q_d(h)=P_d(h)/F_d(h)$ is the proportion of returning indices with a positive target-disease label.

One later admission can label several earlier indices and more than one disease. Positive index--disease pairs, distinct positive indices, distinct positive patients, and distinct outcome admissions are therefore reported separately. A patient-disease event is counted once even when it labels several indices. An outcome admission is counted once across all target diseases. Patient-level counts take the union of positive labels over all eligible index-specific windows. These event counts include first recorded disease admissions that produce at least one positive label within the horizon.

Table~\ref{tab:dataset} reports full-cohort counts before data partitioning. The denominator $N_d$ includes indices without a recorded subsequent admission and is identical at 30, 90, and 365 days.

\paragraph{Cohort summary.}
The candidate cohorts include 33,997 patients in MIMIC-III and 45,195 in MIMIC-IV. Returning candidate indices number 4,418, 7,142, and 9,473 at 30, 90, and 365 days in MIMIC-III, and 9,343, 15,856, and 23,718 in MIMIC-IV. Table~\ref{tab:dataset} reports the admission totals and deduplicated outcomes. Across diseases and horizons, $q_d(h)$ ranges from 6.13\% to 33.16\% in MIMIC-III and 5.34\% to 42.25\% in MIMIC-IV. At 365 days, each returning candidate index has a mean of 1.12 and 0.95 positive disease labels, respectively. At 30 days, $100F_d(30)/N_d$ ranges from 9.15\% to 9.87\% and from 10.79\% to 14.05\%.

Among positive patients, the counts with one, two, three, and at least four target diseases are 2,115, 1,259, 589, and 355 in MIMIC-III and 4,869, 2,924, 1,378, and 870 in MIMIC-IV. The corresponding means are 1.84 and 1.86 diseases per positive patient, and 1.34 and 1.21 positive indices per patient-disease event.

\begin{table}[!htbp]
\caption{Cohort accounting. $N_d$ counts eligible index admissions. Each horizon reports the positive count $P_d(h)$ with its percentage of $N_d$ in parentheses, followed by the count of indices with a completed hospital return $F_d(h)$. The bottom panel reports the deduplicated 365-day totals.}
\label{tab:dataset}
\centering
\footnotesize
\setlength{\tabcolsep}{2.5pt}
\renewcommand{\arraystretch}{1.12}
\begin{tabular*}{\linewidth}{@{\extracolsep{\fill}}lrrrrrrr@{}}
\toprule
\multicolumn{8}{l}{\textit{MIMIC-III}} \\
& & \multicolumn{2}{c}{30 days} & \multicolumn{2}{c}{90 days} & \multicolumn{2}{c}{365 days} \\
\cmidrule(lr){3-4}\cmidrule(lr){5-6}\cmidrule(lr){7-8}
Disease & $N_d$ & $P_d$ (\%) & $F_d$ & $P_d$ (\%) & $F_d$ & $P_d$ (\%) & $F_d$ \\
\midrule
CKD & 35,147 & 524 (1.49) & 3,309 & 962 (2.74) & 5,362 & 1,593 (4.53) & 7,123 \\
HF & 31,173 & 595 (1.91) & 2,853 & 1,129 (3.62) & 4,607 & 1,690 (5.42) & 6,106 \\
COPD & 41,114 & 294 (0.72) & 4,038 & 555 (1.35) & 6,529 & 850 (2.07) & 8,654 \\
HyperK & 38,165 & 585 (1.53) & 3,623 & 1,116 (2.92) & 5,892 & 1,624 (4.26) & 7,796 \\
HypoG & 42,014 & 254 (0.60) & 4,146 & 473 (1.13) & 6,696 & 689 (1.64) & 8,886 \\
Stroke & 40,325 & 266 (0.66) & 3,921 & 518 (1.28) & 6,357 & 820 (2.03) & 8,437 \\
Sepsis & 35,423 & 960 (2.71) & 3,310 & 1,670 (4.71) & 5,286 & 2,316 (6.54) & 6,985 \\
DVT & 41,391 & 449 (1.08) & 4,065 & 763 (1.84) & 6,538 & 1,059 (2.56) & 8,683 \\
\midrule
\multicolumn{8}{l}{\textit{MIMIC-IV}} \\
& & \multicolumn{2}{c}{30 days} & \multicolumn{2}{c}{90 days} & \multicolumn{2}{c}{365 days} \\
\cmidrule(lr){3-4}\cmidrule(lr){5-6}\cmidrule(lr){7-8}
Disease & $N_d$ & $P_d$ (\%) & $F_d$ & $P_d$ (\%) & $F_d$ & $P_d$ (\%) & $F_d$ \\
\midrule
CKD & 39,584 & 726 (1.83) & 4,760 & 1,516 (3.83) & 8,084 & 3,221 (8.14) & 12,161 \\
HF & 36,050 & 766 (2.12) & 4,236 & 1,563 (4.34) & 7,120 & 3,183 (8.83) & 10,722 \\
COPD & 54,504 & 463 (0.85) & 7,517 & 977 (1.79) & 12,797 & 1,927 (3.54) & 19,215 \\
HyperK & 44,466 & 926 (2.08) & 5,492 & 1,842 (4.14) & 9,328 & 3,447 (7.75) & 14,091 \\
HypoG & 56,089 & 421 (0.75) & 7,882 & 876 (1.56) & 13,381 & 1,707 (3.04) & 19,996 \\
Stroke & 51,917 & 501 (0.97) & 7,035 & 1,021 (1.97) & 11,994 & 2,095 (4.04) & 18,021 \\
Sepsis & 36,184 & 1,215 (3.36) & 3,906 & 2,373 (6.56) & 6,572 & 4,232 (11.70) & 10,016 \\
DVT & 53,096 & 829 (1.56) & 7,222 & 1,621 (3.05) & 12,281 & 2,806 (5.28) & 18,394 \\
\midrule
\multicolumn{8}{l}{\textit{Dataset-level counts at 365 days}} \\
\multicolumn{4}{l}{Count} & \multicolumn{2}{r}{MIMIC-III} & \multicolumn{2}{r}{MIMIC-IV} \\
\midrule
\multicolumn{4}{l}{Candidate index admissions $N_0$} & \multicolumn{2}{r}{43,966} & \multicolumn{2}{r}{62,815} \\
\multicolumn{4}{l}{Indices with a completed return $F_0(365)$} & \multicolumn{2}{r}{9,473} & \multicolumn{2}{r}{23,718} \\
\multicolumn{4}{l}{Positive index--disease pairs $\sum_d P_d(365)$} & \multicolumn{2}{r}{10,641} & \multicolumn{2}{r}{22,618} \\
\multicolumn{4}{l}{Distinct positive indices} & \multicolumn{2}{r}{6,116} & \multicolumn{2}{r}{13,486} \\
\multicolumn{4}{l}{Distinct positive patients} & \multicolumn{2}{r}{4,318} & \multicolumn{2}{r}{10,041} \\
\multicolumn{4}{l}{Distinct patient--disease events} & \multicolumn{2}{r}{7,955} & \multicolumn{2}{r}{18,711} \\
\multicolumn{4}{l}{Distinct outcome admissions} & \multicolumn{2}{r}{5,441} & \multicolumn{2}{r}{14,256} \\
\bottomrule
\end{tabular*}
\vspace{3pt}
\end{table}

\subsection{Lead-Time Cohorts}
\label{app:leadtime_cohort}

The seven-day definition removes every index whose first recorded event $t_{i,d}^{\star}$ falls in $(\tau_i,\tau_i+7]$.
These indices are removed instead of relabeled, because a patient diagnosed two days after discharge is not a clean negative at any horizon.
The remaining positives at horizon $h$ are the indices with $t_{i,d}^{\star}\in(\tau_i+7,\tau_i+h]$.
Every method is retrained on the reduced cohorts with the same splits and settings.
Table~\ref{tab:leadtime_counts} reports the counts.
The removed indices are under 1\% of each cohort, but they make up 28\% and 25\% of 30-day positives in MIMIC-III and MIMIC-IV, against 23\% if events were spread evenly over 30 days.
Their share falls to 15\% and 13\% at 90 days and to 10\% and 7\% at 365 days, so the definition mainly changes the 30-day task.

\begin{table}[!htbp]
\caption{Cohort counts under the seven-day lead time. $P_d(7)$ counts indices whose first recorded event falls within seven days of discharge and are removed. $N_d^{(7)}$ is the remaining cohort and $P_h^{(7)}$ its positive count at $h$ days.}
\label{tab:leadtime_counts}
\centering
\footnotesize
\setlength{\tabcolsep}{4pt}
\renewcommand{\arraystretch}{1.08}
\begin{tabular*}{\linewidth}{@{\extracolsep{\fill}}llrrrrr@{}}
\toprule
Data & Disease & $P_d(7)$ & $N_d^{(7)}$ & $P_{30}^{(7)}$ & $P_{90}^{(7)}$ & $P_{365}^{(7)}$ \\
\midrule
MIMIC-III & CKD & 93 & 35,054 & 431 & 869 & 1,500 \\
 & HF & 148 & 31,025 & 447 & 981 & 1,542 \\
 & COPD & 64 & 41,050 & 230 & 491 & 786 \\
 & HyperK & 212 & 37,953 & 373 & 904 & 1,412 \\
 & HypoG & 76 & 41,938 & 178 & 397 & 613 \\
 & Stroke & 51 & 40,274 & 215 & 467 & 769 \\
 & Sepsis & 293 & 35,130 & 667 & 1,377 & 2,023 \\
 & DVT & 152 & 41,239 & 297 & 611 & 907 \\
\midrule
MIMIC-IV & CKD & 118 & 39,466 & 608 & 1,398 & 3,103 \\
 & HF & 169 & 35,881 & 597 & 1,394 & 3,014 \\
 & COPD & 91 & 54,413 & 372 & 886 & 1,836 \\
 & HyperK & 313 & 44,153 & 613 & 1,529 & 3,134 \\
 & HypoG & 119 & 55,970 & 302 & 757 & 1,588 \\
 & Stroke & 103 & 51,814 & 398 & 918 & 1,992 \\
 & Sepsis & 346 & 35,838 & 869 & 2,027 & 3,886 \\
 & DVT & 220 & 52,876 & 609 & 1,401 & 2,586 \\
\bottomrule
\end{tabular*}
\end{table}

\subsection{External Site}
\label{app:ehrshot_protocol}

EHRSHOT~\citep{wornow2023ehrshot} contains longitudinal records of 6,739 Stanford Medicine patients in the OMOP common data model, covering care beyond the ICU.
Its six new-diagnosis tasks ask, at discharge from an inpatient visit, whether a first diagnosis of hypertension, hyperlipidemia, acute myocardial infarction, pancreatic cancer, celiac disease, or lupus will be recorded within one year.
We use the patient-level splits and task labels distributed with the public EHRSHOT benchmark release.
All methods use diagnosis concepts recorded up to the prediction time.
Patient inputs exclude laboratory values, medications, procedures,
and clinical notes.
Diagnosis concepts are mapped to PrimeKG through UMLS, and an evidence graph is built for each target with the same PrimeKG snapshot and construction settings.
The reasoning policy and prompts are reused, and only the EHR encoder and prediction heads are trained on the EHRSHOT training split.
Med-BERT, VoG-adapt, and KARE-adapt follow the same protocol.
Positive rates in the test split range from about 13\% for hypertension and hyperlipidemia to under 1\% for celiac disease and lupus (Table~\ref{tab:ehrshot_counts}).
Appendix~\ref{app:external} reports the results.

\begin{table}[!htbp]
\caption{Test split of the public EHRSHOT benchmark release. A patient can contribute more than one label, and the rate is the share of positive labels.}
\label{tab:ehrshot_counts}
\centering
\small
\setlength{\tabcolsep}{7pt}
\renewcommand{\arraystretch}{1.08}
\begin{tabular}{@{}lrrrr@{}}
\toprule
Task & Patients & Labels & Positive & Rate (\%) \\
\midrule
Hypertension & 755 & 1,258 & 159 & 12.64 \\
Hyperlipidemia & 864 & 1,317 & 172 & 13.06 \\
Acute myocardial infarction & 1,235 & 2,127 & 144 & 6.77 \\
Pancreatic cancer & 1,246 & 2,220 & 56 & 2.52 \\
Celiac disease & 1,255 & 2,222 & 21 & 0.95 \\
Lupus & 1,249 & 2,243 & 20 & 0.89 \\
\bottomrule
\end{tabular}
\end{table}

\subsection{Evidence Graphs}

Table~\ref{tab:kg} summarizes the refined evidence graphs. Appendix~\ref{app:knowledge_preparation} details support scoring and expansion-guide construction.

\begin{table}[!htbp]
\caption{Evidence graphs refined from PrimeKG~\citep{chandak2023building} ($\sim$129K nodes and 4.0M edges).}
\label{tab:kg}
\centering
\footnotesize
\setlength{\tabcolsep}{3pt}
\renewcommand{\arraystretch}{1.12}
\begin{tabular*}{\linewidth}{@{\extracolsep{\fill}}lrrrrr@{}}
\toprule
Target disease & Nodes & Edges & \shortstack{Support\\removed (\%)} & \shortstack{Equivalence\\reduction (\%)} & \shortstack{Guide\\coverage (\%)} \\
\midrule
Chronic kidney disease & 6,841 & 18,273 & 42.3 & 13.7 & 87.4 \\
Heart failure          & 7,214 & 19,847 & 41.8 & 14.2 & 89.1 \\
COPD exacerbation      & 4,923 & 13,516 & 46.1 & 11.3 & 84.6 \\
Hyperkalemia           & 3,847 & 10,284 & 48.7 & 10.8 & 82.3 \\
Hypoglycemia           & 3,612 & 9,738  & 49.2 & 9.6  & 80.7 \\
Ischemic stroke        & 5,731 & 16,142 & 43.6 & 12.9 & 86.8 \\
Sepsis                 & 6,183 & 17,421 & 44.1 & 13.1 & 88.2 \\
Deep vein thrombosis   & 3,294 & 12,847 & 45.8 & 11.4 & 85.3 \\
\midrule
\textit{Average}       & \textit{5,206} & \textit{14,759} & \textit{45.2} & \textit{12.1} & \textit{85.6} \\
\bottomrule
\end{tabular*}
\end{table}

\FloatBarrier
\section{Implementation Details and Pseudocode}
\label{app:impl}

Table~\ref{tab:hyperparams} summarizes the configuration. Algorithm~\ref{alg:reasoning} gives the inference procedure.

\begin{table}[!htbp]
\caption{Hyperparameters and configuration.}
\label{tab:hyperparams}
\centering
\footnotesize
\setlength{\tabcolsep}{3pt}
\renewcommand{\arraystretch}{1.12}
\begin{tabular*}{\linewidth}{@{\extracolsep{\fill}}lrlr@{}}
\toprule
\multicolumn{2}{c}{\textit{Evidence Graph}} & \multicolumn{2}{c}{\textit{Reasoning Loop}} \\
\cmidrule(lr){1-2} \cmidrule(lr){3-4}
Hop limit $K$ & 3 & Budget $B_{\min}$ / $B_{\max}$ & 2 / 12 \\
Degree cap $d_{\max}$ & 50 & Query / expand / select cost & 1.0 / 0.5 / 0.5 \\
Hub edge cap $M$ & 10 & Pre-filter cap $K_{\mathrm{pre}}$ & 5 \\
Support threshold $\tau_s$ & 0.3 & Citation cap $K_C$ & 5 \\
Min edges $K_{\min}$ & 2 & Stop $\epsilon_{\mathrm{stop}}$ / patience $P_{\mathrm{stop}}$ & 0.01 / 2 \\
Quality threshold $\tau_{\mathrm{fb}}$ & 0.4 & Evidence cap $K_E$ & 8 \\
& & Temperature $\beta$ & 2.0 \\
\midrule
\multicolumn{2}{c}{\textit{Model}} & \multicolumn{2}{c}{\textit{Training}} \\
\cmidrule(lr){1-2} \cmidrule(lr){3-4}
Default LLM & GPT-4o-mini & Stage-1 epochs & 30 \\
Second LLM & Qwen3-8B & Stage-2 / stage-3 rounds & 10 / 20 \\
EHR encoder & 2-layer GRU & Episodes per round & $\leq 4{,}096$ \\
Dims ($m$, hidden) & 256, 256 & Predictor / policy LR & $10^{-4}$ / $10^{-5}$ \\
Fusion $f(\cdot)$ & 2-layer MLP & Batch size & 32 \\
& & $(\lambda_{\mathrm{gain}}, \lambda_{\mathrm{cost}}, \lambda_{\mathrm{qual}})$ & (1.0, 0.3, 0.5) \\
& & Hardware & 3$\times$L40S (48GB each) \\
\bottomrule
\end{tabular*}
\end{table}

\subsection{Patient Representations and Knowledge Preparation}
\label{app:knowledge_preparation}

\paragraph{Patient encoding.}
We represent each admission by the mean of its ICD code embeddings, which have dimension 256 and use Xavier uniform initialization. The visits in $H_i$ are ordered chronologically, with the index admission $X_i$ appended as the final visit. A two-layer GRU encodes this sequence, and its final hidden state gives $\mathbf z_i^{\mathrm{ehr}}\in\mathbb R^{256}$. For patients without prior admissions, the sequence contains only the index admission.

\paragraph{ICD-to-KG linking.}
We normalize code punctuation while preserving leading zeros and the ICD version. Each code is joined to UMLS \texttt{MRCONSO.RRF} through the corresponding \texttt{CODE} and \texttt{SAB} fields, using \texttt{ICD9CM} or \texttt{ICD10CM}, to obtain distinct CUIs~\citep{bodenreider2004unified}. Both join sides share this normalization.

PrimeKG's \texttt{umls.py} and \texttt{map\_umls\_mondo.py} workflow produces the CUI-to-MONDO table \texttt{umls\_mondo.csv}~\citep{chandak2023building}. We join its MONDO identifiers to disease nodes in the selected PrimeKG export, respecting that export's grouping of MONDO concepts. Duplicate paths to the same node are collapsed. A code becomes an anchor only when it resolves to one distinct graph node. Unmapped and ambiguous codes are omitted from the anchor set. The resulting $\operatorname{link}(c)$ is single-valued and fixed during training. Other node types remain available during graph traversal.

\paragraph{Edge representation.}
For each edge $e = (u, r, v) \in E_d$, the learned representation $\mathbf{v}(e) \in \mathbb{R}^{256}$ is computed as $\mathbf{v}(e) = \mathrm{MLP}_{\mathrm{edge}}([\mathbf{e}_u;\, \mathbf{e}_r;\, \mathbf{e}_v])$, where $\mathbf{e}_u, \mathbf{e}_v \in \mathbb{R}^{256}$ are the concept node embeddings from Section~\ref{sec:quality_score}, $\mathbf{e}_r \in \mathbb{R}^{256}$ is a learnable relation-type embedding, and $\mathrm{MLP}_{\mathrm{edge}}$ is a single-layer projection from $\mathbb{R}^{768}$ to $\mathbb{R}^{256}$ with ReLU activation.

\paragraph{Knowledge sources.}
\label{app:source_versions}
We use fixed UMLS, MONDO, PrimeKG, and PubMed snapshots across all partitions. The resource manifest records versions, the PrimeKG checksum, mapping-script revision, and corpus date. Each edge stores its source identifier, version, and record key, plus PMIDs and dates for supporting abstracts. Anchor frequencies use only training records. This evaluation restricts EHR inputs to discharge but uses fixed external knowledge. Resource dates are not compared with patient-shifted admission dates.

\paragraph{Evidence support.}
The score $s(e)\in[0,1]$ combines curated-source support with literature alignment. A source table distinguishes expert-curated records from computationally derived associations in PrimeKG's 20 constituent resources~\citep{chandak2023building}. We count distinct curated resources linked to an edge, excluding duplicate exports and reverse copies. Missing source attribution contributes zero. The curated score is
\begin{equation}
s_{\mathrm{curated}}(e)=\min\!\left(1,\frac{n_{\mathrm{curated}}(e)}{3}\right).
\end{equation}
Across the evidence graphs, 91.4\% of edges have $n_{\mathrm{curated}}\leq2$.

For literature support, we query the fixed PubMed abstract corpus with BM25 using the concatenated endpoint concept names. We retrieve the top three abstracts and compute PubMedBERT~\citep{gu2021domain} cosine similarities between the concept-name pair and sentences in these abstracts. Similarities are clipped to $[0,1]$ and averaged to obtain $s_{\mathrm{text}}(e)$. The final score is
\begin{equation}
s(e)=0.4\,s_{\mathrm{curated}}(e)+0.6\,s_{\mathrm{text}}(e).
\end{equation}
Both weights remain fixed. Appendix~\ref{app:sensitivity} reports sensitivity to the curated weight $w_c$ and threshold $\tau_s$.

\paragraph{Expansion guide.}
For disease $d$, an anchor is frequent when its ICD code appears in at least 1\% of training admissions. The guide $I_d$ stores its top 20 incident edge identifiers ranked by $s(e)$. At inference, \textsc{QueryGuide} follows Section~\ref{sec:navigation}, with $K_{\mathrm{pre}}=5$ candidates passed to the LLM.

\subsection{Inference and Budget Accounting}

\begin{algorithm}[!htbp]
\small
\caption{Budget-aware reasoning loop.}
\label{alg:reasoning}
\begin{algorithmic}[1]
\REQUIRE Patient EHR $(X_i,H_i)$, target $(d,h)$,
evidence graph $G_d$,
EHR-only score $\hat{y}_{i,d,h}^{\mathrm{ehr}}$,
max budget $B_{\max}$
\ENSURE Risk score $\hat{y}_{i,d}^{(h)}$, citation list $\mathcal{C}$
 
\STATE $A_{i,d} \leftarrow \{\mathrm{link}(c) : c \in X_i,\; \mathrm{link}(c) \in V_d\}$ \COMMENT{anchor set}
\STATE $\mathbf{p}^{(0)} \leftarrow f_{\mathrm{state}}([\mathbf{z}_i^{\mathrm{ehr}};\, \mathbf{u}(d,h)])$;\quad $E \leftarrow \emptyset$;\quad $C \leftarrow 0$
\STATE $u_i \leftarrow F_{d,h}^{\mathrm{train}}
\!\left(1-\left|2\hat{y}_{i,d,h}^{\mathrm{ehr}}-1\right|\right)$;
\quad $B_i \leftarrow B_{\min}+(B_{\max}-B_{\min})u_i$
\STATE $P = [s_1, \ldots, s_n] \leftarrow \textsc{LLM-Plan}(X_i, H_i, d, h, A_{i,d}, B_i)$ \COMMENT{plan generation}
 
\FOR{each next step $s_t$ in the current plan $P$}
  \IF{the next acquisition and selection attempt is not affordable}
    \STATE \textbf{break}
  \ENDIF
  \STATE Resolve $q_t$ to affordable query actions $\mathcal Q_t$ \COMMENT{query policy, Appendix~\ref{app:routing_policy}}
  \IF{$\mathcal Q_t=\emptyset$}
    \STATE \textbf{break}
  \ENDIF
  \STATE $a_t \leftarrow \arg\max_{a\in\mathcal Q_t}\pi_\theta(a\mid S_t)$; retrieve candidates using $a_t$
  \STATE Score: $\phi_t(e) \leftarrow s(e) \cdot \sigma(\beta \cdot \max\{\mathbf{p}^{(t-1)} \cdot \mathbf{e}_u,\, \mathbf{p}^{(t-1)} \cdot \mathbf{e}_v\})$;\enspace keep top-$K_{\mathrm{pre}}$, remove seen groups
  \STATE $E_t \leftarrow \textsc{LLM-Select}(\text{candidates}, K_E-|E|)$;\enspace $C \leftarrow C + c_t$
  \STATE $\mathrm{result} \leftarrow \textsc{LLM-Verify}(s_t, E_t, \hat{f}_t)$ \COMMENT{verification}
  \IF{$\mathrm{result} = \texttt{FAIL}$}
    \STATE $\mathbf{p}^{(t)} \leftarrow \mathbf{p}^{(t-1)}$
    \IF{a further acquisition and selection attempt is affordable}
      \STATE $P_{t+1:} \leftarrow \textsc{LLM-Revise}(s_t, E_t, E, B_i{-}C)$
    \ENDIF
  \ELSE
    \STATE $E \leftarrow E \cup E_t$;\enspace $\mathbf{p}^{(t)} \leftarrow \mathrm{GRU}(\mathbf{p}^{(t-1)},\, \bar{\mathbf{v}}^{(t)})$
  \ENDIF
  \IF{no further acquisition and selection attempt is affordable \textbf{or} $|E| \geq K_E$}
    \STATE \textbf{break} \COMMENT{budget exhausted or evidence limit}
  \ELSIF{$\Delta U_t < \epsilon_{\mathrm{stop}}$ for $P_{\mathrm{stop}}$ consecutive steps}
    \STATE \textbf{break} \COMMENT{quality stagnation}
  \ENDIF
\ENDFOR
 
\IF{$|E| < K_{\min}$ \textbf{or} $U_{\mathrm{final}} < \tau_{\mathrm{fb}}$}
  \RETURN $\hat{y}_{i,d,h}^{\mathrm{ehr}},\emptyset$ \COMMENT{frozen EHR prediction and no citations}
\ELSE
  \STATE $\mathbf{z}^{\mathrm{kg}} \leftarrow \sum_{e \in E} w(e)\,\mathbf{v}(e)$
\ENDIF
\STATE $\hat{y}_{i,d}^{(h)} \leftarrow
\sigma\!\big(
\mathbf{t}_{d,h}^{\top}
f([\mathbf{z}_i^{\mathrm{ehr}};\,\mathbf{z}^{\mathrm{kg}}])
+b_{d,h}
\big)$
\STATE $\mathcal{C} \leftarrow \text{top-}K_C \text{ edges in } E \text{ by } \phi^\star\!(e)\text{, dedup by } g(e)$
\RETURN $\hat{y}_{i,d}^{(h)},\; \mathcal{C}$
\end{algorithmic}
\end{algorithm}

\paragraph{Inference accounting.}
Table~\ref{tab:latency} counts planning, selection, verification, revision, and retry calls per test episode. Every episode includes an initial plan, and ToG-adapt has no verification calls. A reasoning step is one executed acquisition attempt, including failures. The first attempt uses \textsc{QueryGuide} plus selection and costs 1.5 units. Later attempts use \textsc{QueryExpand} plus selection and cost 1.0 unit each. Empty selections and failed verification remain charged. Zero-attempt episodes incur no acquisition cost. Planning, verification, and revision count as LLM calls, while transport retries add calls without another acquisition charge. Affordability is checked before each attempt.

Budgets follow Eq.~\eqref{eq:adaptive_budget} using the task-specific training uncertainty CDF, held fixed for validation and testing. On the evaluated episodes, $\overline B=2+10\overline u$. Within each run, Budget Utilization is $100\sum_i C_i/\sum_i B_i$, including assigned budgets for zero-attempt episodes. Table~\ref{tab:main} reports its mean and sample SD across runs. The metric is undefined for unbudgeted baselines.

For category $g$, let $B_g=\sum_{i\in g}B_i$ and $U_g=100\sum_{i\in g}C_i/B_g$. Overall utilization is $\sum_g B_gU_g/\sum_g B_g$. These weights reduce to episode-count weights when categories have equal mean assigned budgets.

\begin{table}[!htbp]
\caption{LLM calls and acquisition steps per test episode on MIMIC-III, averaged over three runs. Totals use unrounded counts.}
\label{tab:latency}
\centering
\small
\setlength{\tabcolsep}{3pt}
\renewcommand{\arraystretch}{1.12}
\begin{tabular*}{\linewidth}{@{\extracolsep{\fill}}lrrrrrrr@{}}
\toprule
Method & Plan & Select & Verify & Revise & Retry & Total & Steps \\
\midrule
ToG-adapt & 1.00 & 6.65 & 0.00 & 0.00 & 0.25 & 7.90 & 6.65 \\
VoG-adapt & 1.00 & 5.57 & 5.57 & 0.70 & 0.21 & 13.04 & 5.57 \\
\modelname{} & 1.00 & 3.83 & 3.83 & 0.56 & 0.15 & 9.37 & 3.83 \\
\modelname{} (Qwen3) & 1.00 & 4.38 & 4.38 & 0.79 & 0.17 & 10.73 & 4.38 \\
\bottomrule
\end{tabular*}
\end{table}

Table~\ref{tab:tokens} counts tokens over all LLM calls in an episode.
\modelname{} sends far fewer input tokens than ToG-adapt or VoG-adapt and more output tokens, because its plans and verifications are longer while each step shows only $K_{\mathrm{pre}}$ filtered candidates.
The ratios are nearly the same on both datasets.

\begin{table}[!htbp]
\caption{Mean input, output, and total LLM tokens per test episode at the default setting.}
\label{tab:tokens}
\centering
\small
\setlength{\tabcolsep}{4pt}
\renewcommand{\arraystretch}{1.12}
\begin{tabular*}{\linewidth}{@{\extracolsep{\fill}}lrrrrrr@{}}
\toprule
& \multicolumn{3}{c}{MIMIC-III} & \multicolumn{3}{c}{MIMIC-IV} \\
\cmidrule(lr){2-4}\cmidrule(lr){5-7}
Method & Input & Output & Total & Input & Output & Total \\
\midrule
ToG-adapt & 13,841 & 1,517 & 15,358 & 15,541 & 1,607 & 17,148 \\
VoG-adapt & 18,420 & 1,954 & 20,374 & 21,076 & 2,101 & 23,177 \\
\modelname{} & 7,935 & 2,341 & 10,276 & 8,873 & 2,452 & 11,325 \\
\bottomrule
\end{tabular*}
\end{table}

Most episodes include at least three acquisition attempts (Table~\ref{tab:step_distribution}).

\begin{table}[!htbp]
\caption{Reasoning-step distribution (\%), pooling episodes across three runs.}
\label{tab:step_distribution}
\centering
\small
\setlength{\tabcolsep}{8pt}
\renewcommand{\arraystretch}{1.12}
\begin{tabular}{@{}llrrr@{}}
\toprule
Dataset & Method & 1 step (\%) & 2 steps (\%) & $\geq3$ steps (\%) \\
\midrule
MIMIC-III & \modelname{} & 7.47 & 23.43 & 69.11 \\
MIMIC-III & \modelname{} (Qwen3) & 7.45 & 26.59 & 65.96 \\
\midrule
MIMIC-IV & \modelname{} & 5.94 & 25.75 & 68.31 \\
MIMIC-IV & \modelname{} (Qwen3) & 6.21 & 26.77 & 67.02 \\
\bottomrule
\end{tabular}
\end{table}

\subsection{Training}

\paragraph{Initialization and sampling.}
Learnable embeddings use Xavier uniform initialization.
The LLM uses temperature 0.0 for plan generation and verification,
and 0.7 for exploration during training.
Stage~1 uses all eligible training indices for 30 epochs.
Stages~2 and~3 use 10 and 20 sampling rounds, respectively.
Each round draws up to 4,096 eligible training indices per dataset
uniformly without replacement, with a new draw in each round.
Each selected index contributes one episode, with a disease sampled
uniformly from its eligible disease set and a horizon sampled uniformly
from $\{30,90,365\}$ days.
The stage~2 draws are shared across LLM-based methods within each seed.
Stages~2 and~3 retain the final checkpoint of each stage. For each completed configuration and seed, full validation predictions are generated once and reused for hyperparameter selection and recalibration. 

\paragraph{Evaluation calls.}
Offline API evaluation uses Batch requests, with dependent reasoning steps submitted after their preceding outputs are available. Responses to deterministic calls are reused only when the model version, prompt, complete inputs, and decoding settings match. Training exploration uses fresh calls, and stage~3 collects fresh on-policy trajectories.

\paragraph{Policy optimization details.}
\label{app:routing_policy}
The routing policy learns which concrete graph query to execute. The LLM generates and revises the plan, selects returned edges, and verifies evidence. Its weights stay frozen. At step $t$, the routing state $S_t$ contains the current patient state $\mathbf p_{i,d,h}^{(t-1)}$, accepted evidence, current query intent $q_t$, target $(d,h)$, and remaining budget $B_i-C_{t-1}$.

The query intent is resolved to allowed node and relation identifiers in $G_d$. The initial action set contains \textsc{QueryGuide} calls on allowed anchors. Later sets contain \textsc{QueryExpand}$(x,r)$ calls on allowed frontier nodes and incident relation types. A frontier node is an anchor or an endpoint of accepted evidence. Invalid identifiers, queries with no unused admissible edges, and actions whose query-plus-selection cost exceeds the remaining budget are masked out. Let $\mathcal Q_t$ denote the remaining admissible action set. An empty action set ends the episode. Each selected action is one acquisition attempt.

A two-layer MLP with a 256-unit ReLU hidden layer assigns a scalar logit
\begin{equation}
\ell_\theta(S_t,a)=\mathrm{MLP}_{\theta}\!\left(
[\mathbf p_{i,d,h}^{(t-1)};\mathbf e_x;\mathbf e_r;\mathbf u(d,h);(B_i-C_{t-1})/B_i]
\right).
\label{eq:router_logit}
\end{equation}
For \textsc{QueryGuide}, $x$ is the anchor and the relation feature is zero. For \textsc{QueryExpand}, $x$ and $r$ identify the frontier node and relation. The action distribution is
\begin{equation}
\pi_\theta(a\mid S_t)=
\frac{\exp\ell_\theta(S_t,a)}
{\sum_{a'\in\mathcal Q_t}\exp\ell_\theta(S_t,a')},
\qquad a\in\mathcal Q_t.
\label{eq:router_policy}
\end{equation}
Stage~2 uses a uniform distribution over the same admissible action sets. Stage~3 samples from $\pi_\theta$, and evaluation selects the highest-probability action, with ties resolved by stable node and relation identifiers. After a query, candidate scoring, top-$K_{\mathrm{pre}}$ filtering, LLM edge selection, and verification follow Section~\ref{sec:navigation}.

Here $\theta$ contains the routing MLP parameters. The EHR encoder, graph embeddings, patient-state update, and predictor have parameters $\psi$ and receive supervised updates. Within each policy batch, $\psi$ and all LLM outputs are held fixed for the routing update. The rollout stores the log probability of every sampled action, including failed attempts. For a batch of $n$ episodes, the policy loss is
\begin{equation}
\mathcal L_{\mathrm{policy}}(\theta)
=-\frac{1}{n}\sum_{i=1}^{n}
\operatorname{stopgrad}(r_{i,d_i,h_i}-b_k)
\sum_{t=1}^{T_i}\log\pi_\theta(a_{i,t}\mid S_{i,t}).
\label{eq:policy_loss}
\end{equation}
The return is the undiscounted episode reward in Eq.~\eqref{eq:reward}. The baseline $b_k$ is the moving average from earlier batches, initialized to zero and updated after the current policy step as $b_{k+1}=0.99b_k+0.01\bar r_k$, where $\bar r_k$ is the mean episode reward in batch $k$. The routing update uses the policy learning rate in Table~\ref{tab:hyperparams}. A separate binary cross-entropy update uses the predictor learning rate for $\psi$. Gradients from the policy loss do not pass through the reward, discrete graph lookups, or LLM calls.

\subsection{Evaluation Protocol}

\paragraph{Aggregation and uncertainty.}
\label{app:run_allocation}
Results are averaged over three independent training seeds,
with the same seeds and data splits used across methods.
Within each run, AUPRC is macro-averaged over diseases and horizons.
Table~\ref{tab:main} also reports sample SDs.
Per-disease, per-horizon, efficiency, and clinical-utility summaries
reuse the main runs.
Auxiliary experiments use matched reference configurations,
with gains and drops computed within each seed and dataset before averaging.
Identical configurations reuse stored predictions within each seed.
EHRSHOT intervals use paired patient bootstrap with trained models held fixed.
All aggregates use unrounded values.

Citation precision uses the sampled citations described in
Appendix~\ref{app:citation_eval}.
Evidence quality pools all returned citations within each run.
Fallback episodes contribute to prediction, fallback, and coverage statistics.
Coverage is the fraction of episodes with citations and equals
one minus the fallback rate.
Category and overall fallback rates pool episode counts within
each run before averaging across seeds.

\paragraph{Baseline adaptation.}
\label{app:baseline_fairness}
EHR-only baselines (RETAIN, StageNet, Med-BERT) receive the same ICD codes and prediction targets.
Static KG baselines (GRAM, GraphCare) use PrimeKG~\citep{chandak2023building} as their knowledge source.
GRAM learns attention-based code representations using the disease hierarchy extracted from PrimeKG.
GraphCare constructs personalized patient-KG subgraphs from PrimeKG and applies graph neural networks to obtain patient representations.
Both use the full PrimeKG rather than the refined evidence graph $G_d$, following their original designs.

ToG-adapt and VoG-adapt share the refined evidence graph $G_d$, GPT-4o-mini, and the ICD-to-KG linker with \modelname{}.
ToG-adapt follows the iterative beam search of Think-on-Graph 2.0~\citep{ma2025thinkongraph}, where the LLM generates queries, retrieves candidate edges, and selects among them for multiple rounds.
VoG-adapt follows the stepwise verification protocol of VoG~\citep{zhao2026vog}, where the LLM verifies each retrieved edge against its hypothesis before proceeding.
Neither method imposes a budget constraint, so both run until the LLM produces a stop signal or a maximum of 10 reasoning steps is reached.
Table~\ref{tab:latency} reports their call and step counts.
Neither method uses support-score differentiation during acquisition or citation-integrity training.
Both use the same EHR representation and predictor as \modelname{} (Section~\ref{sec:predictor}), with uniform edge pooling in place of quality-weighted pooling.

KARE-adapt follows KARE~\citep{jiang2025reasoningenhanced}, which groups a medical KG into hierarchical communities, summarizes each community with an LLM, and retrieves the summaries most relevant to a patient.
We run its community detection and summarization on the same evidence graph $G_d$ with GPT-4o-mini and retrieve summaries for the concepts that the shared ICD-to-KG linker returns.
The retrieved summaries feed the same EHR representation and predictor head as the other KG-augmented methods.
This replaces KARE's original multi-source graph and fine-tuned LLM predictor, so the comparison isolates the retrieval strategy.
KARE-adapt has no acquisition budget, and its citations are community summaries.

All baselines use the same binary cross-entropy loss and patient-disjoint splits.
EHR-only and static KG baselines use the full training cohorts for
30 and 40 epochs, respectively.
ToG-adapt, VoG-adapt, and KARE-adapt use 30 epochs of EHR-only warm-up
followed by the same 10 sampled supervised rounds as stage~2 of \modelname{}.

\FloatBarrier
\section{Detailed Results}
\label{app:full_results}

\subsection{Per-Disease and Per-Horizon Results}
\label{app:per_task}

\paragraph{Per-disease performance.}

\modelname{} has the highest AUPRC on every disease except sepsis, where KARE-adapt is slightly ahead on both datasets (Table~\ref{tab:per_disease}).
The gains vary within each disease category, and no single baseline leads across tasks.

\begin{table}[!htbp]
\caption{AUPRC (\%) per disease, averaged over three horizons within each run and then over three runs. Avg matches the mean reported in Table~\ref{tab:main}. Bold marks the best value in each column.}
\label{tab:per_disease}
\centering
\small
\setlength{\tabcolsep}{3pt}
\renewcommand{\arraystretch}{1.12}
\begin{tabular*}{\linewidth}{@{\extracolsep{\fill}}lrrrrrrrrr@{}}
\toprule
Method & CKD & HF & COPD & HyperK & HypoG & Stroke & Sepsis & DVT & Avg \\
\midrule\multicolumn{10}{l}{\textit{MIMIC-III}} \\
RETAIN & 10.69 & 12.05 & 8.84 & 12.00 & 8.58 & 6.02 & 11.43 & 6.64 & 9.53 \\
StageNet & 12.08 & 12.79 & 9.90 & 13.71 & 8.99 & 6.62 & 12.98 & 8.00 & 10.63 \\
Med-BERT & 13.42 & 13.70 & 10.53 & 14.71 & 10.57 & 6.82 & 13.33 & 8.42 & 11.44 \\
GRAM & 14.52 & 14.89 & 11.71 & 15.41 & 10.83 & 8.08 & 14.39 & 9.77 & 12.45 \\
GraphCare & 16.44 & 17.14 & 13.08 & 17.41 & 12.41 & 9.26 & 16.15 & 12.44 & 14.29 \\
ToG-adapt & 18.06 & 19.71 & 14.30 & 19.83 & 14.07 & 10.94 & 18.46 & 14.21 & 16.20 \\
VoG-adapt & 17.97 & 18.70 & 14.82 & 18.23 & 13.89 & 10.72 & 18.71 & 14.51 & 15.94 \\
KARE-adapt & 16.22 & 14.80 & 16.43 & 19.81 & 14.59 & 12.20 & \textbf{22.09} & 12.31 & 16.06 \\
\modelname{} & \textbf{22.56} & \textbf{23.11} & \textbf{17.92} & \textbf{23.34} & \textbf{16.92} & \textbf{13.55} & 22.07 & \textbf{16.55} & \textbf{19.50} \\
\midrule\multicolumn{10}{l}{\textit{MIMIC-IV}} \\
RETAIN & 10.20 & 11.52 & 7.90 & 12.09 & 8.28 & 5.50 & 11.20 & 6.31 & 9.13 \\
StageNet & 11.78 & 11.83 & 8.10 & 12.29 & 8.46 & 6.12 & 12.01 & 6.68 & 9.66 \\
Med-BERT & 12.99 & 13.43 & 10.63 & 14.22 & 9.23 & 6.23 & 12.80 & 8.83 & 11.04 \\
GRAM & 16.10 & 16.35 & 11.05 & 15.15 & 9.66 & 6.97 & 13.70 & 10.70 & 12.46 \\
GraphCare & 15.45 & 16.52 & 11.97 & 16.87 & 12.23 & 8.20 & 15.25 & 13.04 & 13.69 \\
ToG-adapt & 16.41 & 17.03 & 13.36 & 18.09 & 13.27 & 9.42 & 17.04 & 11.86 & 14.56 \\
VoG-adapt & 18.07 & 18.81 & 13.55 & 17.19 & 14.30 & 9.63 & 17.46 & 14.28 & 15.41 \\
KARE-adapt & 14.98 & 18.29 & 14.49 & 19.99 & 15.16 & 9.31 & \textbf{22.40} & 12.45 & 15.88 \\
\modelname{} & \textbf{21.20} & \textbf{22.41} & \textbf{16.33} & \textbf{22.05} & \textbf{16.32} & \textbf{12.62} & 22.00 & \textbf{16.64} & \textbf{18.70} \\
\bottomrule
\end{tabular*}
\end{table}

\begin{figure}[!htbp]
\centering
\includegraphics[width=0.78\linewidth]{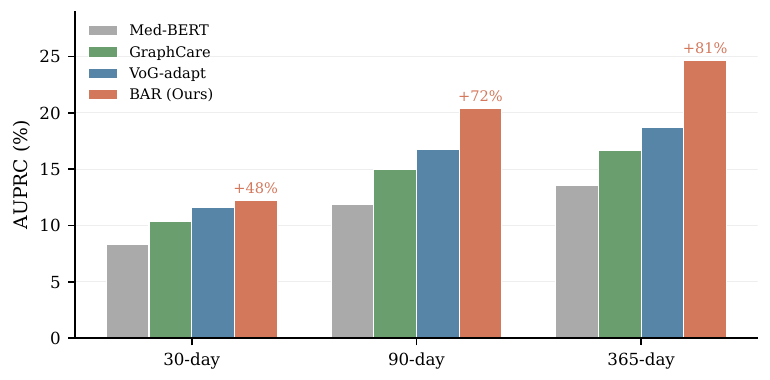}
\caption{Per-horizon AUPRC averaged across both datasets. Percentages show the relative gain of \modelname{} over the best EHR-only baseline.}
\label{fig:per_horizon}
\end{figure}

\paragraph{Prediction horizon.}
The relative gain over the strongest EHR-only baseline grows with the prediction horizon (Figure~\ref{fig:per_horizon}).

\paragraph{Fallback and coverage.}
Fallback is more frequent for drug-disease interactions than for comorbidity cascades (Table~\ref{tab:fallback_coverage}).
Citation coverage is lower on MIMIC-IV, which falls back more often.

\begin{table}[!htbp]
\caption{Fallback rates and citation coverage (\%). Category rates pool episode counts within each run before averaging runs.}
\label{tab:fallback_coverage}
\centering
\small
\setlength{\tabcolsep}{10pt}
\renewcommand{\arraystretch}{1.12}
\begin{tabular}{@{}lrr@{}}
\toprule
 & MIMIC-III & MIMIC-IV \\
\midrule
\multicolumn{3}{l}{\textit{Fallback rate (\%)}} \\
Comorbidity cascades & 11.42 & 15.46 \\
Disease progression & 14.11 & 18.40 \\
Drug-disease interactions & 22.58 & 26.84 \\
\midrule
Overall fallback (\%) & 15.54 & 19.75 \\
Citation coverage (\%) & 84.46 & 80.25 \\
\bottomrule
\end{tabular}
\end{table}

\subsection{Results under the Seven-Day Lead Time}
\label{app:leadtime_results}

Every method keeps 76--85\% of its primary macro AUPRC under the lead-time definition, and the EHR-only, static KG, Graph4LLM, and \modelname{} groups keep their AUPRC order (Table~\ref{tab:leadtime_main}).
KARE-adapt has the highest baseline AUPRC on both datasets, narrowly ahead of ToG-adapt on MIMIC-III and VoG-adapt on MIMIC-IV, while VoG-adapt has the highest baseline AUROC.
Against the baselines, \modelname{} leads on seven of eight diseases on MIMIC-III and six on MIMIC-IV (Table~\ref{tab:leadtime_disease}).
It ties KARE-adapt on COPD in MIMIC-IV and trails it on sepsis on both datasets, as under the primary definition.
By horizon, the margin over the strongest comparator stays within 0.4 points at 30 days, slightly negative on MIMIC-IV, and grows at 90 and 365 days (Table~\ref{tab:leadtime_horizon}).
The 30-day task loses about a quarter of its positives to the lead time, and it is also the horizon where external evidence adds least under the primary definition.

\begin{table}[!htbp]
\caption{Macro AUROC and AUPRC (\%) under the seven-day lead-time definition.}
\label{tab:leadtime_main}
\centering
\small
\setlength{\tabcolsep}{8pt}
\renewcommand{\arraystretch}{1.08}
\begin{tabular}{@{}llrrrr@{}}
\toprule
& & \multicolumn{2}{c}{MIMIC-III} & \multicolumn{2}{c}{MIMIC-IV} \\
\cmidrule(lr){3-4}\cmidrule(lr){5-6}
Category & Method & AUROC & AUPRC & AUROC & AUPRC \\
\midrule
EHR-only & RETAIN & 69.84 & 7.60 & 68.90 & 7.49 \\
 & StageNet & 71.79 & 8.77 & 70.78 & 7.90 \\
 & Med-BERT & 73.03 & 9.09 & 72.51 & 9.03 \\
Static KG & GRAM & 72.71 & 9.42 & 73.45 & 9.53 \\
 & GraphCare & 76.15 & 11.71 & 75.52 & 11.26 \\
Graph4LLM & ToG-adapt & 77.30 & 13.20 & 76.28 & 11.92 \\
 & VoG-adapt & 77.76 & 12.65 & 77.56 & 12.95 \\
 & KARE-adapt & 77.39 & 13.34 & 76.14 & 12.97 \\
\modelname{} & Qwen3 & 79.39 & 14.16 & 78.15 & 14.28 \\
 & GPT-4o-mini & \textbf{80.94} & \textbf{15.75} & \textbf{80.42} & \textbf{15.43} \\
\bottomrule
\end{tabular}
\end{table}

\begin{table}[!htbp]
\caption{AUPRC (\%) by disease under the seven-day lead time, averaged over the three horizons. Bold marks the best value in each column.}
\label{tab:leadtime_disease}
\centering
\scriptsize
\setlength{\tabcolsep}{3pt}
\renewcommand{\arraystretch}{1.08}
\begin{tabular*}{\linewidth}{@{\extracolsep{\fill}}lrrrrrrrr@{}}
\toprule
Method & CKD & HF & COPD & HyperK & HypoG & Stroke & Sepsis & DVT \\
\midrule
\multicolumn{9}{l}{MIMIC-III} \\
RETAIN & 8.71 & 9.15 & 8.14 & 9.32 & 6.74 & 5.27 & 8.96 & 4.54 \\
StageNet & 10.40 & 10.67 & 8.40 & 11.88 & 7.53 & 5.74 & 10.12 & 5.39 \\
Med-BERT & 11.96 & 10.85 & 8.83 & 10.42 & 8.21 & 5.57 & 10.54 & 6.32 \\
GRAM & 12.74 & 10.92 & 9.70 & 10.69 & 7.51 & 6.33 & 10.36 & 7.11 \\
GraphCare & 14.51 & 15.27 & 11.12 & 14.18 & 9.39 & 7.28 & 12.33 & 9.60 \\
ToG-adapt & 14.86 & 16.64 & 12.78 & 14.93 & 10.71 & 10.39 & 14.68 & 10.59 \\
VoG-adapt & 14.97 & 14.51 & 12.45 & 13.39 & 10.69 & 9.96 & 14.70 & 10.56 \\
KARE-adapt & 14.23 & 12.06 & 14.10 & 16.46 & 12.20 & 10.68 & 18.05 & 8.97 \\
\modelname{} (Qwen3) & 17.23 & 16.20 & 14.21 & \textbf{16.56} & 11.77 & 9.01 & \textbf{18.26} & 10.03 \\
\modelname{} & \textbf{20.34} & \textbf{19.91} & \textbf{15.66} & 16.54 & \textbf{13.30} & \textbf{10.89} & 16.96 & \textbf{12.39} \\
\midrule
\multicolumn{9}{l}{MIMIC-IV} \\
RETAIN & 9.22 & 9.71 & 6.42 & 8.62 & 7.08 & 4.85 & 9.08 & 4.95 \\
StageNet & 10.49 & 10.01 & 6.88 & 8.37 & 7.22 & 4.95 & 9.80 & 5.44 \\
Med-BERT & 10.99 & 11.22 & 9.08 & 11.40 & 7.07 & 5.23 & 10.53 & 6.77 \\
GRAM & 11.57 & 12.60 & 9.00 & 11.94 & 7.15 & 5.67 & 10.60 & 7.73 \\
GraphCare & 13.22 & 14.17 & 10.77 & 12.66 & 10.19 & 7.27 & 11.70 & 10.09 \\
ToG-adapt & 14.41 & 14.05 & 11.37 & 13.40 & 10.61 & 8.08 & 14.02 & 9.45 \\
VoG-adapt & 16.76 & 14.99 & 11.72 & 13.51 & 12.01 & 8.13 & 13.86 & 12.60 \\
KARE-adapt & 12.89 & 16.04 & 12.55 & 14.49 & 11.24 & 7.78 & \textbf{18.74} & 10.02 \\
\modelname{} (Qwen3) & \textbf{19.38} & 16.76 & \textbf{13.51} & 16.10 & 11.96 & 9.20 & 15.88 & 11.49 \\
\modelname{} & 19.29 & \textbf{19.57} & 12.55 & \textbf{17.81} & \textbf{12.77} & \textbf{10.28} & 18.12 & \textbf{13.06} \\
\bottomrule
\end{tabular*}
\end{table}

\begin{table}[!htbp]
\caption{AUPRC (\%) by horizon under the seven-day lead time, against the strongest non-\modelname{} comparator at each horizon. Differences are percentage points.}
\label{tab:leadtime_horizon}
\centering
\small
\setlength{\tabcolsep}{7pt}
\renewcommand{\arraystretch}{1.08}
\begin{tabular}{@{}lrlrrr@{}}
\toprule
Data & $h$ & Comparator & Comparator AUPRC & \modelname{} AUPRC & Difference \\
\midrule
MIMIC-III & 30 & VoG-adapt & 7.77 & 8.13 & 0.36 \\
 & 90 & KARE-adapt & 14.09 & 17.04 & 2.95 \\
 & 365 & KARE-adapt & 18.64 & 22.07 & 3.43 \\
MIMIC-IV & 30 & VoG-adapt & 7.79 & 7.48 & $-$0.31 \\
 & 90 & VoG-adapt & 14.08 & 16.74 & 2.66 \\
 & 365 & KARE-adapt & 18.42 & 22.07 & 3.65 \\
\bottomrule
\end{tabular}
\end{table}

\FloatBarrier
\section{Robustness and Generalization}
\label{app:robustness}

\subsection{Navigation Policy and Candidate Corruption}
\label{app:navigation}

To isolate the role of the LLM in navigation, we replace plan-guided navigation with two fixed policies at the same assigned budgets.
The support-greedy policy expands the candidate with the highest support score, and the random policy picks uniformly among admissible candidates.
Support scoring, verification, and the budget rule are unchanged.
Both policies lose more AUPRC than removing verification from \modelname{}, and random navigation falls below GraphCare (Table~\ref{tab:navigation}).
The support-greedy policy cites relevant edges more often than the loop without verification, which suggests that well-attested edges tend to be relevant even when they add little to the prediction.

To test the loop under noisy retrieval, we replace 25\% or 50\% of the returned candidates at every navigation step on MIMIC-IV with irrelevant edges sampled from $G_d$, keeping the candidate count and the assigned budget fixed.
AUPRC declines linearly with the replaced share, while citation precision declines faster at the higher level and fallback absorbs a growing share of episodes (Table~\ref{tab:corruption}).
Verification therefore turns many corrupted steps into fallback, but some irrelevant edges still reach the citation list.
Even with half of the candidates replaced, \modelname{} stays well above the Stage-1 EHR model that it falls back to.

\begin{table}[!htbp]
\centering
\small
\begin{minipage}[t]{0.54\linewidth}
\centering
\caption{AUPRC and citation precision (Cit.) of navigation policies at matched assigned budgets (\%).}
\label{tab:navigation}
\setlength{\tabcolsep}{4pt}
\renewcommand{\arraystretch}{1.08}
\begin{tabular}{@{}lrrrr@{}}
\toprule
& \multicolumn{2}{c}{MIMIC-III} & \multicolumn{2}{c}{MIMIC-IV} \\
\cmidrule(lr){2-3}\cmidrule(lr){4-5}
Policy & AUPRC & Cit. & AUPRC & Cit. \\
\midrule
\modelname{} & \textbf{19.50} & \textbf{77.06} & \textbf{18.70} & \textbf{73.64} \\
No verification & 16.75 & 61.18 & 15.99 & 56.92 \\
Support-greedy & 15.79 & 65.44 & 15.46 & 59.96 \\
Random & 13.89 & 59.45 & 13.01 & 52.07 \\
\bottomrule
\end{tabular}
\end{minipage}\hfill
\begin{minipage}[t]{0.42\linewidth}
\centering
\caption{Candidate replacement on MIMIC-IV (\%). Stage-1 EHR is the fallback predictor.}
\label{tab:corruption}
\setlength{\tabcolsep}{4pt}
\renewcommand{\arraystretch}{1.08}
\begin{tabular}{@{}lrrr@{}}
\toprule
Replaced & AUPRC & Fallback & Cit. \\
\midrule
None & 18.70 & 19.75 & 73.64 \\
25\% & 17.50 & 25.90 & 63.65 \\
50\% & 16.26 & 34.09 & 50.09 \\
Stage-1 EHR & 11.17 & --- & --- \\
\bottomrule
\end{tabular}
\end{minipage}
\end{table}

\subsection{Input Sparsity and Disease Transfer}

\begin{figure}[!htbp]
\centering
\includegraphics[width=\linewidth]{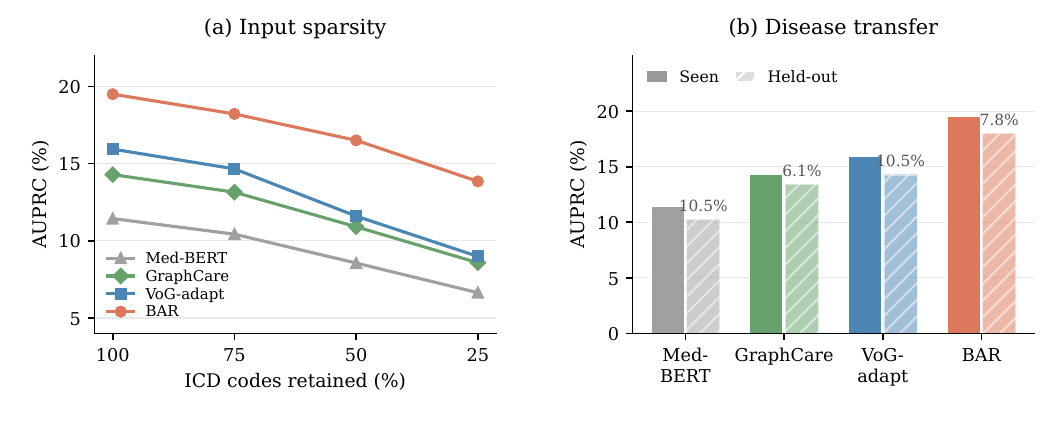}
\caption{Robustness on MIMIC-III. \textbf{(a)} AUPRC when ICD codes are randomly dropped at test time without retraining. \textbf{(b)} Leave-two-out disease transfer, where the pipeline is trained on six diseases and only the prediction heads are fine-tuned on the two held-out diseases. Percentages show the AUPRC decline from seen to held-out diseases.}
\label{fig:robustness}
\end{figure}

AUPRC decreases as fewer ICD codes are retained (Figure~\ref{fig:robustness}a). \modelname{} maintains the highest AUPRC across the retention levels, and its relative advantage over VoG-adapt grows as codes are removed.

\modelname{} has the highest AUPRC on held-out diseases (Figure~\ref{fig:robustness}b). Its relative decline is 7.8\%, compared with 6.1\% for GraphCare and 10.5\% for both Med-BERT and VoG-adapt.

\subsection{Knowledge Source and Linker Coverage}
\label{app:coverage}

To reduce linker coverage, we randomly remove 25\% or 50\% of the code types that the linker maps, rebuild the anchor sets and evidence graphs, and retrain every KG method.
To change the knowledge source, we replace PrimeKG with a graph built from UMLS relations~\citep{bodenreider2004unified} and rerun graph construction with the same settings.
Both methods lose AUPRC as linker coverage falls, while \modelname{} remains ahead in every setting (Table~\ref{tab:coverage}).
The relative margin varies across coverage levels and knowledge sources.

A code type has a unique match when it resolves to exactly one graph node (Appendix~\ref{app:knowledge_preparation}).
Unmatched rates are highest for external-cause, supplementary, and injury codes, which describe circumstances, encounters, or injuries that seldom have a disease node in the graph (Table~\ref{tab:mapping}).
Among disease and symptom codes, 62\% of MIMIC-III types and 64\% and 55\% of MIMIC-IV ICD-9-CM and ICD-10-CM types have a unique match.
ICD-10-CM adds laterality and encounter detail to many codes, which may explain why fewer of its disease codes resolve to a single node.

\begin{table}[!htbp]
\centering
\footnotesize
\begin{minipage}[t]{0.45\linewidth}
\centering
\caption{AUPRC (\%) under reduced linker coverage and under UMLS, against the strongest baseline in Table~\ref{tab:main}.}
\label{tab:coverage}
\setlength{\tabcolsep}{3.5pt}
\renewcommand{\arraystretch}{1.08}
\begin{tabular}{@{}lrrrrr@{}}
\toprule
& & \multicolumn{2}{c}{MIMIC-III} & \multicolumn{2}{c}{MIMIC-IV} \\
\cmidrule(lr){3-4}\cmidrule(lr){5-6}
Graph & Codes & \modelname{} & ToG & \modelname{} & KARE \\
\midrule
PrimeKG & 100\% & 19.50 & 16.20 & 18.70 & 15.88 \\
 & 75\% & 17.49 & 15.18 & 16.71 & 14.08 \\
 & 50\% & 15.65 & 13.27 & 14.90 & 12.47 \\
UMLS & 100\% & 18.97 & 15.87 & 17.88 & 15.33 \\
\bottomrule
\end{tabular}
\end{minipage}\hfill
\begin{minipage}[t]{0.51\linewidth}
\centering
\caption{Code types without a unique graph match, out of all code types. Supplementary codes are ICD-9-CM V or ICD-10-CM Z codes.}
\label{tab:mapping}
\setlength{\tabcolsep}{3.5pt}
\renewcommand{\arraystretch}{1.08}
\begin{tabular}{@{}lrrr@{}}
\toprule
& MIMIC-III & \multicolumn{2}{c}{MIMIC-IV} \\
\cmidrule(lr){2-2}\cmidrule(lr){3-4}
Code group & ICD-9 & ICD-9 & ICD-10 \\
\midrule
Disease, symptom & 1,446/3,841 & 1,578/4,390 & 4,548/10,183 \\
Injury, poisoning & 650/838 & 962/1,219 & 4,359/4,733 \\
Supplementary & 489/579 & 436/507 & 1,873/2,067 \\
External causes & 231/235 & 303/306 & 683/709 \\
\bottomrule
\end{tabular}
\end{minipage}
\end{table}

\subsection{External Site}
\label{app:external}

Among the evaluated methods, \modelname{} achieves the highest AUPRC
on five of six EHRSHOT tasks and trails KARE-adapt and Med-BERT
on lupus (Table~\ref{tab:external}).
Its macro AUPRC is 13.74, compared with 10.45 for VoG-adapt,
the strongest baseline by macro AUPRC.
Against KARE-adapt, the paired 95\% intervals exclude zero for
hyperlipidemia, acute myocardial infarction, and celiac disease.
Hyperlipidemia accounts for most of the macro gain over KARE-adapt.
Excluding this task, the macro AUPRC is 10.13 for \modelname{}
and 9.36 for KARE-adapt, with a difference of 0.78 points
calculated before rounding.

\begin{table}[!htbp]
\caption{AUPRC (\%) on EHRSHOT. Differences from KARE-adapt and their paired 95\% bootstrap intervals are in percentage points. Table~\ref{tab:ehrshot_counts} gives the test counts.}
\label{tab:external}
\centering
\small
\setlength{\tabcolsep}{5pt}
\renewcommand{\arraystretch}{1.08}
\begin{tabular}{@{}lrrrrr@{}}
\toprule
Task & Med-BERT & VoG-adapt & KARE-adapt & \modelname{} & Difference [95\% CI] \\
\midrule
Hypertension & 16.78 & 18.42 & 19.85 & \textbf{20.50} & 0.64 [$-$1.07, 2.55] \\
Hyperlipidemia & 18.60 & 26.49 & 14.85 & \textbf{31.78} & 16.93 [13.51, 20.86] \\
Acute myocardial infarction & 14.00 & 8.27 & 14.29 & \textbf{16.17} & 1.88 [0.10, 3.81] \\
Pancreatic cancer & 3.80 & 5.04 & 5.52 & \textbf{6.06} & 0.54 [$-$0.68, 2.20] \\
Celiac disease & 2.53 & 2.03 & 2.44 & \textbf{3.82} & 1.37 [0.08, 5.66] \\
Lupus & 4.33 & 2.45 & \textbf{4.68} & 4.12 & $-$0.56 [$-$2.81, 3.14] \\
\midrule
Macro & 10.00 & 10.45 & 10.27 & \textbf{13.74} & --- \\
\bottomrule
\end{tabular}
\end{table}

\subsection{Laboratory and Medication Inputs}
\label{app:labs}

On MIMIC-IV, we add the last laboratory values recorded before discharge and the medications active at discharge to the patient input of every method, with the same splits and training budget.
Every method improves (Table~\ref{tab:labs}).
KARE-adapt has the largest gain, and EHR-only methods do not gain consistently more than graph-based ones.
\modelname{} keeps the highest AUROC and AUPRC, with a 1.3-point AUPRC margin over KARE-adapt.

\begin{table}[!htbp]
\caption{Results on MIMIC-IV with laboratory and medication inputs. AUROC and AUPRC are percentages, and the gain is the AUPRC change from diagnosis-only inputs in percentage points.}
\label{tab:labs}
\centering
\small
\setlength{\tabcolsep}{6pt}
\renewcommand{\arraystretch}{1.08}
\begin{tabular}{@{}lrrr@{}}
\toprule
Method & AUROC & AUPRC & AUPRC gain \\
\midrule
RETAIN & 74.04 & 11.64 & 2.51 \\
StageNet & 75.75 & 11.87 & 2.21 \\
Med-BERT & 76.19 & 12.62 & 1.58 \\
GRAM & 77.58 & 13.78 & 1.32 \\
GraphCare & 79.10 & 16.13 & 2.44 \\
ToG-adapt & 79.73 & 16.89 & 2.33 \\
VoG-adapt & 79.93 & 17.01 & 1.60 \\
KARE-adapt & 80.02 & 18.90 & 3.02 \\
\modelname{} (Qwen3) & 80.48 & 18.48 & 1.63 \\
\modelname{} & \textbf{82.87} & \textbf{20.18} & 1.48 \\
\bottomrule
\end{tabular}
\end{table}

\section{Clinical Utility}
\label{app:utility}

For each model, we fit a monotone recalibration map on the validation split and apply it unchanged to the test split.
The Brier score and the integrated calibration index~(ICI)~\citep{austin2019integrated} are computed per task on the probability scale and macro-averaged.
ICI measures the mean absolute difference between predicted probabilities and a smoothed calibration curve.
Net benefit at threshold $t$ is $\mathrm{TP}/n-(\mathrm{FP}/n)\,t/(1-t)$~\citep{vickers2006decision}, where $n$ is the number of test observations.
We report net benefit per 1,000 observations, macro-averaged over the eight diseases.
Positive decisions represent referral for follow-up review.
Reviewing every patient is the reference strategy, and reviewing none has zero net benefit.
For each horizon, the three thresholds sit below, near, and above the macro-averaged test prevalence.

The raw Stage-1 Brier scores exceed the task-wise constant
test-prevalence reference on both datasets.
Validation-based recalibration reduces the Brier score and ICI
for all three models.
\modelname{} has the lowest Brier score before and after
recalibration (Table~\ref{tab:calibration}).
Decision curves use the recalibrated probabilities.
All three models achieve higher net benefit than reviewing
every patient at the evaluated thresholds, and \modelname{}
has the highest net benefit across all 18 settings
(Table~\ref{tab:dca}).

\begin{table}[!htbp]
\caption{Brier score and ICI before and after validation-only recalibration, on the probability scale.}
\label{tab:calibration}
\centering
\small
\setlength{\tabcolsep}{9pt}
\renewcommand{\arraystretch}{1.08}
\begin{tabular}{@{}llrrrr@{}}
\toprule
& & \multicolumn{2}{c}{Brier} & \multicolumn{2}{c}{ICI} \\
\cmidrule(lr){3-4}\cmidrule(lr){5-6}
Data & Model & Before & After & Before & After \\
\midrule
MIMIC-III & Stage-1 EHR & 0.0260 & 0.0223 & 0.0271 & 0.0058 \\
 & KARE-adapt & 0.0248 & 0.0217 & 0.0272 & 0.0060 \\
 & \modelname{} & 0.0238 & \textbf{0.0209} & 0.0273 & 0.0058 \\
MIMIC-IV & Stage-1 EHR & 0.0427 & 0.0366 & 0.0335 & 0.0088 \\
 & KARE-adapt & 0.0397 & 0.0352 & 0.0326 & 0.0088 \\
 & \modelname{} & 0.0381 & \textbf{0.0344} & 0.0328 & 0.0089 \\
\bottomrule
\end{tabular}
\end{table}

\begin{table}[!htbp]
\caption{Decision curves after recalibration. Test prevalence $\pi$ and thresholds $t$ are percentages, and net benefit is per 1,000 test observations. Horizons $h$ are in days.}
\label{tab:dca}
\centering
\small
\setlength{\tabcolsep}{5pt}
\renewcommand{\arraystretch}{1.05}
\begin{tabular*}{\linewidth}{@{\extracolsep{\fill}}lrrrrrrr@{}}
\toprule
Data & $h$ & $\pi$ & $t$ & Stage-1 EHR & KARE-adapt & \modelname{} & Review all \\
\midrule
MIMIC-III & 30 & 1.32 & 0.5 & 8.29 & 8.80 & \textbf{9.29} & 8.23 \\
 & & & 1.5 & 4.16 & 5.33 & \textbf{6.23} & $-$1.84 \\
 & & & 3.0 & 2.37 & 3.48 & \textbf{4.22} & $-$17.33 \\
 & 90 & 2.40 & 1.0 & 15.28 & 16.14 & \textbf{16.86} & 14.17 \\
 & & & 2.5 & 9.23 & 10.74 & \textbf{12.69} & $-$1.00 \\
 & & & 5.0 & 4.86 & 6.72 & \textbf{8.96} & $-$27.34 \\
 & 365 & 3.53 & 1.0 & 26.12 & 26.69 & \textbf{27.58} & 25.52 \\
 & & & 4.0 & 13.49 & 15.79 & \textbf{17.74} & $-$4.93 \\
 & & & 8.0 & 6.53 & 9.19 & \textbf{11.89} & $-$48.62 \\
\midrule
MIMIC-IV & 30 & 1.80 & 0.5 & 13.14 & 13.22 & \textbf{13.85} & 13.07 \\
 & & & 2.0 & 6.40 & 7.26 & \textbf{8.46} & $-$2.03 \\
 & & & 4.0 & 3.21 & 4.27 & \textbf{5.48} & $-$22.91 \\
 & 90 & 3.58 & 1.0 & 26.56 & 27.01 & \textbf{28.10} & 26.06 \\
 & & & 3.5 & 14.28 & 16.16 & \textbf{18.91} & 0.83 \\
 & & & 8.0 & 6.10 & 8.50 & \textbf{11.45} & $-$48.05 \\
 & 365 & 6.75 & 2.0 & 50.45 & 51.82 & \textbf{53.21} & 48.44 \\
 & & & 6.5 & 26.68 & 32.94 & \textbf{36.34} & 2.64 \\
 & & & 12.0 & 11.07 & 19.37 & \textbf{24.05} & $-$59.69 \\
\bottomrule
\end{tabular*}
\end{table}

\section{Hyperparameter Sensitivity}
\label{app:sensitivity}

\subsection{Reasoning Hyperparameters}

\begin{figure}[!htbp]
\centering
\includegraphics[width=\linewidth]{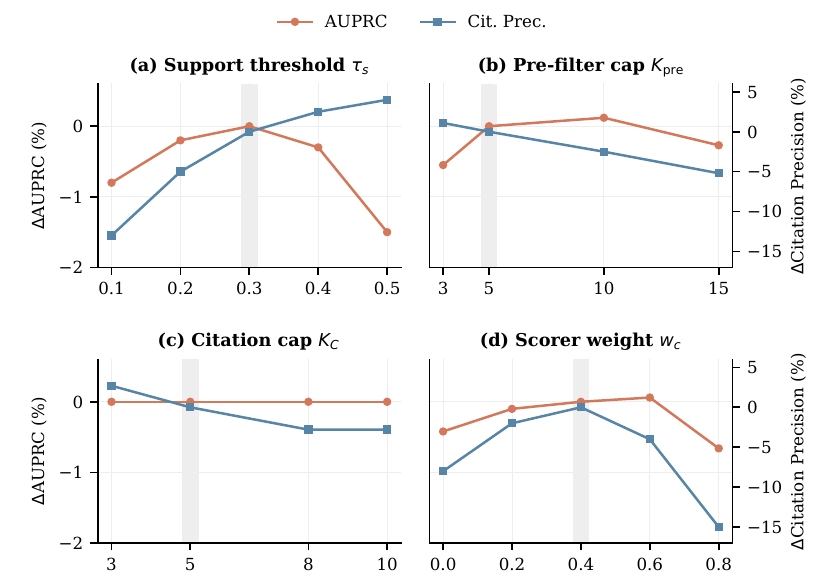}
\caption{Sensitivity to key hyperparameters, averaged across both datasets. Changes are relative to the default setting in each panel. Gray bands mark defaults.}
\label{fig:sensitivity}
\end{figure}

Figure~\ref{fig:sensitivity} shows how prediction and citation quality change around the default settings.
Lowering the support threshold below its default admits weaker evidence and reduces both metrics. Raising it improves citation precision but lowers AUPRC, suggesting that useful evidence is also excluded.
The default $\tau_s=0.3$ balances these effects.

Increasing the pre-filter cap offers little prediction benefit and lowers citation precision. We retain $K_{\mathrm{pre}}=5$ to keep the candidate list short.
At inference, $K_C$ limits the final citation list after prediction. Returning more edges lowers citation precision while leaving AUPRC unchanged. The curves level off once $K_C$ reaches the evidence limit $K_E=8$.

AUPRC changes little across moderate values of the scorer weight, while citation precision is highest at the default $w_c=0.4$.
Both metrics decline when either the textual or curated signal receives most of the weight, supporting their combined use.

\subsection{Budget Ceiling}
\label{app:budget_sweep}

Table~\ref{tab:budget_sweep} lists the sweep behind Figure~\ref{fig:budget} for each dataset.
Token use grows quickly up to $B_{\max}=8$ and slowly afterward, which fits the stopping rule and the evidence cap ending many episodes before the budget is spent, as the 62--65\% utilization at the default also shows.

\begin{table}[!htbp]
\caption{Budget sweep. Tokens are mean LLM tokens per episode, and AUPRC is a percentage. The default is $B_{\max}=12$.}
\label{tab:budget_sweep}
\centering
\footnotesize
\setlength{\tabcolsep}{3pt}
\renewcommand{\arraystretch}{1.1}
\begin{tabular*}{\linewidth}{@{\extracolsep{\fill}}llrrrrrrrrrr@{}}
\toprule
\multicolumn{2}{@{}l}{$B_{\max}$} & 2 & 4 & 6 & 8 & 10 & 12 & 14 & 16 & 18 & 20 \\
\midrule
MIMIC-III & Tokens & 3,677 & 5,674 & 7,484 & 8,957 & 9,857 & 10,276 & 10,673 & 11,047 & 11,373 & 11,655 \\
 & AUPRC & 15.78 & 18.67 & 19.68 & \textbf{19.72} & 19.62 & 19.50 & 19.38 & 19.28 & 19.18 & 19.10 \\
MIMIC-IV & Tokens & 3,916 & 6,043 & 8,015 & 9,696 & 10,806 & 11,325 & 11,691 & 12,028 & 12,308 & 12,547 \\
 & AUPRC & 14.87 & 17.56 & 18.62 & \textbf{18.78} & 18.77 & 18.70 & 18.62 & 18.54 & 18.47 & 18.41 \\
\bottomrule
\end{tabular*}
\end{table}

\subsection{Graph Construction}
\label{app:graph_sweeps}
 
We vary $K$, $d_{\max}$, $M$, and anchor frequency one at a time.
For each dataset, the default graph and four alternatives start independently from each seed's default checkpoint.
Each receives five additional stage~3 sampling rounds of up to 4,096 episodes, with matched patient--disease--horizon draws across settings.
Routing and supervised updates follow Appendix~\ref{app:routing_policy}.
Embeddings are copied by node and relation identifier, and newly introduced embeddings use Xavier initialization.
Training samples all three horizons, while validation and testing for this analysis use 365 days.
Each setting retains its fifth-round checkpoint.

On MIMIC-IV, one-shot retrieval, support-greedy navigation, removal of verification, and removal of plan revision are applied at inference to each setting's checkpoint.
Each variant reruns the reasoning loop with model parameters fixed.
Drops are measured against the full model from the same checkpoint.

Full-model AUPRC stays within 0.6 points of the default and above the strongest baselines at this horizon, ToG-adapt on MIMIC-III (20.87) and KARE-adapt on MIMIC-IV (21.16).
The ablations keep their ranking except at $M=20$, where support-greedy navigation costs slightly more than one-shot retrieval.
 
\begin{table}[!htbp]
\caption{Graph-construction sensitivity at 365 days after five rounds of adaptation. Default denotes the adapted default graph with $K=3$, $d_{\max}=50$, $M=10$, and anchor frequency 1\%. Full-model AUPRC is a percentage. MIMIC-IV ablation drops are percentage-point differences from the same row's full model, with model parameters fixed during ablation.}
\label{tab:graph_sweeps}
\centering
\small
\setlength{\tabcolsep}{3.5pt}
\renewcommand{\arraystretch}{1.08}
\begin{tabular}{@{}lrrrrrr@{}}
\toprule
& \multicolumn{2}{c}{Full model} & \multicolumn{4}{c}{AUPRC drop on MIMIC-IV} \\
\cmidrule(lr){2-3}\cmidrule(lr){4-7}
Setting & MIMIC-III & MIMIC-IV & One-shot & Support-greedy & w/o verification & w/o revision \\
\midrule
Default & 24.91 & 24.35 & 4.52 & 3.68 & 3.09 & 1.43 \\
$K=2$ & 24.38 & 24.09 & 3.76 & 3.34 & 2.63 & 1.12 \\
$d_{\max}=100$ & 24.72 & 23.98 & 4.41 & 3.95 & 3.42 & 1.59 \\
$M=20$ & 25.06 & 24.58 & 4.18 & 4.29 & 3.17 & 1.36 \\
Anchor freq.\ 0.5\% & 24.67 & 24.22 & 4.67 & 3.52 & 2.88 & 1.54 \\
\bottomrule
\end{tabular}
\end{table}

\section{Evidence-Quality Score Analysis}
\label{app:quality_analysis}

Table~\ref{tab:quadrants} groups observations by median support and patient relevance. The Pearson correlation between the scores is 0.228.
Selection is highest when both scores are high, lowest when both are low, and similar in the two mixed groups.

\begin{table}[!htbp]
\caption{Selection rates (\%) and selected/total counts for 40,000 observations, split at the median of each score. Marginal rates pool the counts.}
\label{tab:quadrants}
\centering
\footnotesize
\setlength{\tabcolsep}{10pt}
\renewcommand{\arraystretch}{1.12}
\begin{tabular}{@{}lrrr@{}}
\toprule
 & High Relevance & Low Relevance & Marginal Rate \\
\midrule
High Support
 & 66.5 (7,699/11,584)
 & 30.5 (2,571/8,416)
 & 51.4 \\
Low Support
 & 31.1 (2,620/8,416)
 & 16.5 (1,913/11,584)
 & 22.7 \\
\midrule
Marginal Rate & 51.6 & 22.4 & --- \\
\bottomrule
\end{tabular}
\end{table}

\section{Citation Precision Evaluation}
\label{app:citation_eval}

Citation precision measures the relevance of returned edges to the patient's risk pathway.
For each method, dataset, seed, and evaluated setting, we sample up to 5,000 returned citation occurrences uniformly without replacement.
All occurrences are used when fewer are available.
Sampling treats repeated citations from different episodes as separate occurrences.
GPT-4o~\citep{hurst2024gpt} receives the discharge codes, target disease, and cited edge with its support score, and returns a binary relevance label with a short explanation at temperature 0~\citep{zheng2023judging}.
The prediction and the edge's position in the reasoning trace are hidden.
Citation precision is the proportion labeled relevant in the sample.
Judgments are reused for identical complete judge inputs.
The 300 doubly annotated pairs below assess judge agreement separately.

\paragraph{Annotation protocol.}
The comparison covers 300 patient-edge pairs. Both annotators label the same pairs independently using the judge's inputs and relevance criterion.

\paragraph{Agreement.}
Table~\ref{tab:citation_agreement} reports pairwise agreement and judge performance on the 254 pairs with a shared annotator label. Appendix~\ref{app:prompts} gives the judge prompt.

\begin{table}[!htbp]
\caption{Pairwise Cohen's $\kappa$ on 300 patient-edge pairs. The lower panel evaluates the judge on the 254 pairs where both annotators agree. Precision and recall refer to the relevant class.}
\label{tab:citation_agreement}
\centering
\small
\setlength{\tabcolsep}{10pt}
\renewcommand{\arraystretch}{1.12}
\begin{tabular}{@{}lrrr@{}}
\toprule
 & Annotator A & Annotator B & GPT-4o \\
\midrule
Annotator A & --- & 0.69 & 0.76 \\
Annotator B & 0.69 & --- & 0.65 \\
GPT-4o & 0.76 & 0.65 & --- \\
\midrule
Judge vs. shared label & $\kappa$ & Precision & Recall \\
Consensus subset & 0.83 & 0.89 & 0.93 \\
\bottomrule
\end{tabular}
\end{table}

\section{LLM Prompt Templates}
\label{app:prompts}

We use five prompt types, one for plan generation, one for edge selection during navigation, one for verification, one for plan revision, and one for the citation judge.
All reasoning prompts receive the patient's discharge codes $X_i$, history $H_i$, target disease $d$, horizon $h$, and remaining budget.

\begin{reasoningprompt}{Plan generation}
\small
\textbf{Instruction:} Decompose a clinical risk assessment into evidence-gathering steps over a medical knowledge graph. Each step traces a specific clinical pathway (comorbidity cascade, drug interaction, or disease progression) from the patient's conditions toward the target disease. Allocate budget fractions that sum to at most 1.0.

\smallskip
\begin{tabular}{@{}l@{\enspace}l@{\qquad}l@{\enspace}l}
\textbf{Inputs} & & & \\
\quad Patient codes & $\{X_i\}$ & History & $\{H_i\}$ \\
\quad Target disease & $d$ & Horizon & $h$ days \\
\quad Anchor set & $A_{i,d}$ & Budget & $B$ \\
\end{tabular}

\smallskip
\textbf{Output:} $P = [s_1, \ldots, s_n]$ where $s_t = (\text{hypothesis } h_t,\; \text{query } q_t,\; \text{expected } \hat{f}_t,\; \text{budget } b_t)$.

\smallskip
{\small Example $s_1$\,: hypothesis ``AF raises clot risk'', query ``AF $\to$ thromboembolism'', expected ``stasis, hypercoagulability'', budget 0.35}
\end{reasoningprompt}

\begin{reasoningprompt}{Edge selection (navigation)}
\small
\textbf{Instruction:} Given the current hypothesis and a ranked list of candidate edges that have already been filtered by support threshold and deduplicated, select which edges to retain as evidence for this step. Prioritize edges that directly support the hypothesis.

\smallskip
\begin{tabular}{@{}l@{\enspace}l}
\textbf{Inputs} & \\
\quad Current step & $s_t = (h_t, q_t, \hat{f}_t, b_t)$ \\
\quad Candidates & $\{(u_j, r_j, v_j, \phi_t(e_j))\}_{j=1}^{K_{\mathrm{pre}}}$, ranked by $\phi_t$ \\
\quad Remaining capacity & $K_E-|E|$ \\
\end{tabular}

\smallskip
\textbf{Output:} $E_t \subseteq \text{candidates}$, with at most $K_E-|E|$ selected edge IDs.
\end{reasoningprompt}

\begin{reasoningprompt}{Verification}
\small
\textbf{Instruction:} Verify whether the retrieved evidence supports the plan step hypothesis. A step passes if the edges are clinically consistent with the hypothesis and the average quality score is sufficient. A step fails if the evidence is absent, contradictory, or too weak to support the hypothesis.

\smallskip
\begin{tabular}{@{}l@{\enspace}l@{\qquad}l@{\enspace}l}
\textbf{Inputs} & & & \\
\quad Hypothesis & $h_t$ & Expected & $\hat{f}_t$ \\
\quad Retrieved & $E_t$ with $\{\phi_t(e)\}$ & & \\
\end{tabular}

\smallskip
\textbf{Output:} \texttt{PASS} $|$ \texttt{FAIL} with one-sentence reason.
\end{reasoningprompt}

\begin{reasoningprompt}{Plan revision}
\small
\looseness=-1 \textbf{Instruction:} Regenerate the plan from the failed step onward using a different clinical pathway that avoids the failure mode. Previously verified steps are frozen. The revised plan must fit within the remaining budget.

\smallskip
\begin{tabular}{@{}l@{\enspace}l}
\textbf{Inputs} & \\
\quad Failed step & $s_t$ with $E_t$ and failure reason \\
\quad Verified context & $\{(s_j, E_j)\}_{j=1}^{t-1}$ \\
\quad Remaining budget & $B - C_t$ \\
\end{tabular}

\smallskip
\textbf{Output:} A revised plan for the remaining steps.
\end{reasoningprompt}

\begin{judgeprompt}{Citation relevance judge}
\small
\textbf{Instruction:} Determine whether a knowledge graph edge cited by a clinical reasoning system is relevant to the patient's risk for the target disease. An edge is relevant if it describes a mechanism, risk factor, or pathway connecting the patient's conditions to the target. 

\smallskip
\begin{tabular}{@{}l@{\enspace}l}
\textbf{Inputs} & \\
\quad Patient codes & $\{X_i\}$ \\
\quad Target disease & $d$ \\
\quad Cited edge & $(u, r, v)$ with support $s(e)$ \\
\end{tabular}

\smallskip
\textbf{Output:} \texttt{RELEVANT} $|$ \texttt{NOT\_RELEVANT}, with one-sentence justification.
\end{judgeprompt}

\section{Broader Impact and Ethics}
\label{app:ethics}

\paragraph{Data use.}
The study design uses de-identified MIMIC-III~\citep{johnson2016mimic} and MIMIC-IV~\citep{johnson2023mimic} records under the PhysioNet Credentialed Health Data License and de-identified EHRSHOT~\citep{wornow2023ehrshot} records under its research data use agreement.
UMLS is used under the UMLS Metathesaurus License.

\paragraph{Intended use and limitations.}
\modelname{} is a research prototype for budget-aware reasoning over medical knowledge graphs. Clinical deployment would require prospective validation and expert review of its reasoning traces, neither of which is included in this study.

\paragraph{Potential risks.}
PrimeKG~\citep{chandak2023building} may be incomplete or outdated, and LLM reasoning may be clinically incorrect.
Follow-up covers only admissions within each health system, with no recorded readmission labeled negative, so diagnoses made elsewhere are unobserved.
Fixed-horizon labels do not model censoring or competing risks such as death.
Raw risk scores are miscalibrated and need local recalibration before use as probabilities.
Citation evaluation checks whether cited edges are relevant to the patient, and the medical correctness of full reasoning traces remains untested.
Generalization to other populations and subgroup differences by age, sex, race, or insurance remain untested.

\FloatBarrier
\stopcontents[appendix]

\section*{NeurIPS Paper Checklist}

\begin{enumerate}

\item {\bf Claims}
    \item[] Question: Do the main claims made in the abstract and introduction accurately reflect the paper's contributions and scope?
    \item[] Answer: \answerYes{}
    \item[] Justification: The abstract and introduction describe the evidence graphs, budget-aware reasoning loop, and training objective. Their empirical claims follow Table~\ref{tab:main}, Figure~\ref{fig:budget}, Figure~\ref{fig:per_horizon}, and Appendix~\ref{app:external}, within the scope stated in Section~\ref{sec:experiments}.

\item {\bf Limitations}
    \item[] Question: Does the paper discuss the limitations of the work performed by the authors?
    \item[] Answer: \answerYes{}
    \item[] Justification: Section~\ref{sec:conclusion} states that labels cover only diagnoses recorded within one health system and do not model censoring or competing risks, that raw scores need local recalibration, that the medical correctness of full reasoning traces is untested, and that prospective clinical validation is still needed. Section~\ref{sec:analysis} reports that the margin over the baselines disappears at 30 days under a seven-day lead time. Tables~\ref{tab:latency} and~\ref{tab:tokens} report the LLM calls and tokens per episode, and Appendix~\ref{app:ethics} discusses further risks, including untested subgroup differences.

\item {\bf Theory assumptions and proofs}
    \item[] Question: For each theoretical result, does the paper provide the full set of assumptions and a complete (and correct) proof?
    \item[] Answer: \answerNA{}
    \item[] Justification: The paper does not include formal theoretical results or proofs. The contributions are methodological and empirical.

    \item {\bf Experimental result reproducibility}
    \item[] Question: Does the paper fully disclose all the information needed to reproduce the main experimental results of the paper to the extent that it affects the main claims and/or conclusions of the paper (regardless of whether the code and data are provided or not)?
    \item[] Answer: \answerYes{}
    \item[] Justification: We specify datasets, target diseases, prediction horizons, and a patient-disjoint split based on recorded dates. The methodology and appendix describe the EHR encoder, support scorer, hyperparameters, training procedure, reasoning-loop pseudocode, baseline adaptations, and the lead-time and EHRSHOT protocols.

\item {\bf Open access to data and code}
    \item[] Question: Does the paper provide open access to the data and code, with sufficient instructions to faithfully reproduce the main experimental results, as described in supplemental material?
    \item[] Answer: \answerYes{}
    \item[] Justification: MIMIC-III and MIMIC-IV are available through PhysioNet credentialed access, EHRSHOT under its research data use agreement, and UMLS under its license, and PrimeKG is publicly available. Code is released at the repository linked in Section~\ref{sec:introduction}. The paper provides sufficient algorithmic detail for independent reimplementation.

\item {\bf Experimental setting/details}
    \item[] Question: Does the paper specify all the training and test details (e.g., data splits, hyperparameters, how they were chosen, type of optimizer) necessary to understand the results?
    \item[] Answer: \answerYes{}
    \item[] Justification: The experiment section describes datasets, baselines, and metrics. The appendix gives the model configuration, patient-disjoint recorded-date split, baseline adaptations, and the details of cohort extraction, linking, and policy optimization.

\item {\bf Experiment statistical significance}
    \item[] Question: Does the paper report error bars suitably and correctly defined or other appropriate information about the statistical significance of the experiments?
    \item[] Answer: \answerYes{}
    \item[] Justification: Table~\ref{tab:main} reports means and sample SDs over three independent training seeds. Table~\ref{tab:external} reports paired patient-bootstrap intervals. Appendix~\ref{app:run_allocation} specifies the run allocation, and Appendix~\ref{app:citation_eval} describes citation sampling.

\item {\bf Experiments compute resources}
    \item[] Question: For each experiment, does the paper provide sufficient information on the computer resources (type of compute workers, memory, time of execution) needed to reproduce the experiments?
    \item[] Answer: \answerNo{}
    \item[] Justification: Table~\ref{tab:hyperparams} reports hardware, and Tables~\ref{tab:latency} and~\ref{tab:tokens} report LLM calls and tokens per test episode. Training times and total compute are not reported.
    
\item {\bf Code of ethics}
    \item[] Question: Does the research conducted in the paper conform, in every respect, with the NeurIPS Code of Ethics \url{https://neurips.cc/public/EthicsGuidelines}?
    \item[] Answer: \answerYes{}
    \item[] Justification: The research uses only de-identified records accessed under their data use terms, namely MIMIC-III and MIMIC-IV under the PhysioNet Credentialed Health Data License and EHRSHOT under its research data use agreement. No re-identification was attempted. Appendix~\ref{app:ethics} discusses data use, intended use, limitations, and potential risks.

\item {\bf Broader impacts}
    \item[] Question: Does the paper discuss both potential positive societal impacts and negative societal impacts of the work performed?
    \item[] Answer: \answerYes{}
    \item[] Justification: Appendix~\ref{app:ethics} discusses potential risks, including outdated KG knowledge, clinically plausible but incorrect LLM reasoning, miscalibrated raw scores, unobserved diagnoses outside the health system, and untested subgroup differences, and it states that clinical use would require prospective validation and expert review of the reasoning traces.
    
\item {\bf Safeguards}
    \item[] Question: Does the paper describe safeguards that have been put in place for responsible release of data or models that have a high risk for misuse (e.g., pre-trained language models, image generators, or scraped datasets)?
    \item[] Answer: \answerNA{}
    \item[] Justification: The paper does not release pre-trained models or scraped datasets. \modelname{} is a research prototype evaluated on existing public benchmarks and poses no direct risk for misuse.

\item {\bf Licenses for existing assets}
    \item[] Question: Are the creators or original owners of assets (e.g., code, data, models), used in the paper, properly credited and are the license and terms of use explicitly mentioned and properly respected?
    \item[] Answer: \answerYes{}
    \item[] Justification: MIMIC-III and MIMIC-IV are cited and used under the PhysioNet Credentialed Health Data License, EHRSHOT under its research data use agreement, and UMLS under the UMLS Metathesaurus License. PrimeKG is cited. The LLMs (GPT-4o-mini, GPT-4o, Qwen3-8B) are cited with their respective references.

\item {\bf New assets}
    \item[] Question: Are new assets introduced in the paper well documented and is the documentation provided alongside the assets?
    \item[] Answer: \answerYes{}
    \item[] Justification: The paper releases its code at the repository linked in Section~\ref{sec:introduction}.

\item {\bf Crowdsourcing and research with human subjects}
    \item[] Question: For crowdsourcing experiments and research with human subjects, does the paper include the full text of instructions given to participants and screenshots, if applicable, as well as details about compensation (if any)? 
    \item[] Answer: \answerNA{}
    \item[] Justification: The paper does not involve crowdsourcing or research with human subjects. The two-annotator validation in Appendix~\ref{app:citation_eval} (300 instances for citation precision) is an internal evaluation step, not a human-subjects study.

\item {\bf Institutional review board (IRB) approvals or equivalent for research with human subjects}
    \item[] Question: Does the paper describe potential risks incurred by study participants, whether such risks were disclosed to the subjects, and whether Institutional Review Board (IRB) approvals (or an equivalent approval/review based on the requirements of your country or institution) were obtained?
    \item[] Answer: \answerNA{}
    \item[] Justification: The paper uses only de-identified records (MIMIC-III, MIMIC-IV, and EHRSHOT) under their existing data use agreements. No human subjects research was conducted.

\item {\bf Declaration of LLM usage}
    \item[] Question: Does the paper describe the usage of LLMs if it is an important, original, or non-standard component of the core methods in this research? Note that if the LLM is used only for writing, editing, or formatting purposes and does \emph{not} impact the core methodology, scientific rigor, or originality of the research, declaration is not required.
    \item[] Answer: \answerYes{}
    \item[] Justification: LLMs (GPT-4o-mini, Qwen3-8B) are core components of the proposed method, serving as the reasoning agent in the budget-aware reasoning loop. Their role is described throughout the Methodology section, with prompt templates in Appendix~\ref{app:prompts}. GPT-4o serves as the citation judge (Appendix~\ref{app:citation_eval}).

\end{enumerate} 

\end{document}